\documentclass[runningheads]{llncs}

\usepackage{eccv}

\usepackage{eccvabbrv}
\usepackage{eso-pic}
\usepackage{graphicx}
\usepackage{float}
\usepackage{booktabs}
\usepackage{makecell}
\usepackage[accsupp]{axessibility}  

\usepackage{hyperref}

\usepackage{orcidlink}

\makeatletter
\renewcommand{\maketitle}{
  \begingroup
  \setlength{\textwidth}{2\textwidth}
  \@maketitle
  \endgroup
}
\makeatother
\begin{document}

\titlerunning{CoralscapesV2}

\author{\small
Jonathan Sauder\inst{1} \and
Thomas Ruckli\inst{2} \and
Gabrielė Strodomskytė\inst{2} \and\\
Ibrahim Souleiman Abdallah\inst{3} \and 
Rahma Hassan Abdi\inst{3} \and
Djama Goumaneh Awaleh\inst{4} \and
Mohamed Houssein Farah\inst{4} \and
Moustapha Nour\inst{5} \and
Osama Sharhubil Saad\inst{6} \and 
Mustafa Mohammed Khalafallah Altaib\inst{6} \and
Maysoon Kteifan\inst{7} \and
Farah Alsoqi\inst{7} \and
Eyad Zgool\inst{7} \and
Jafar Al-Omari\inst{8} \and
Temesgen Gebremeskel Gebreluel\inst{9} \and
Zekaria Zekeria Abdulkerim\inst{9} \and
Meron Ghirmay\inst{9} \and
Teklehaimanot Beraki\inst{10} \and
Devis Tuia\inst{2} \and
Guilhem Banc-Prandi\inst{2}
}

\authorrunning{Sauder et al.}

\institute{\scriptsize MIT
\and EPFL
\and University of Djibouti
\and Ministry of Environment and Sustainable Development of Djibouti
\and Centre d'Études et de Recherche de Djibouti
\and Red Sea University of Port Sudan
\and Aqaba Marine Reserve, Jordan
\and United Nations Development Programme Jordan
\and Department of Biology, Mai Nefhi College of Science, Eritrea
\and Ministry of Marine Resources, Massawa, Eritrea}

\title{CoralscapesV2: Panoptic and Fine-Grained Visual Scene Understanding in Coral Reefs\vspace{-14pt}}

\maketitle
\AddToShipoutPictureFG*{%
  \AtPageLowerLeft{%
    \raisebox{15mm}{%
      \makebox[\paperwidth][c]{%
        \scriptsize ECCV 2026 Workshop on Marine Vision, Malmö, Sweden
      }%
    }%
  }%
}

\begin{center}
    \centering
\includegraphics[trim={0 125px 0 125px},clip, width=0.995\linewidth]{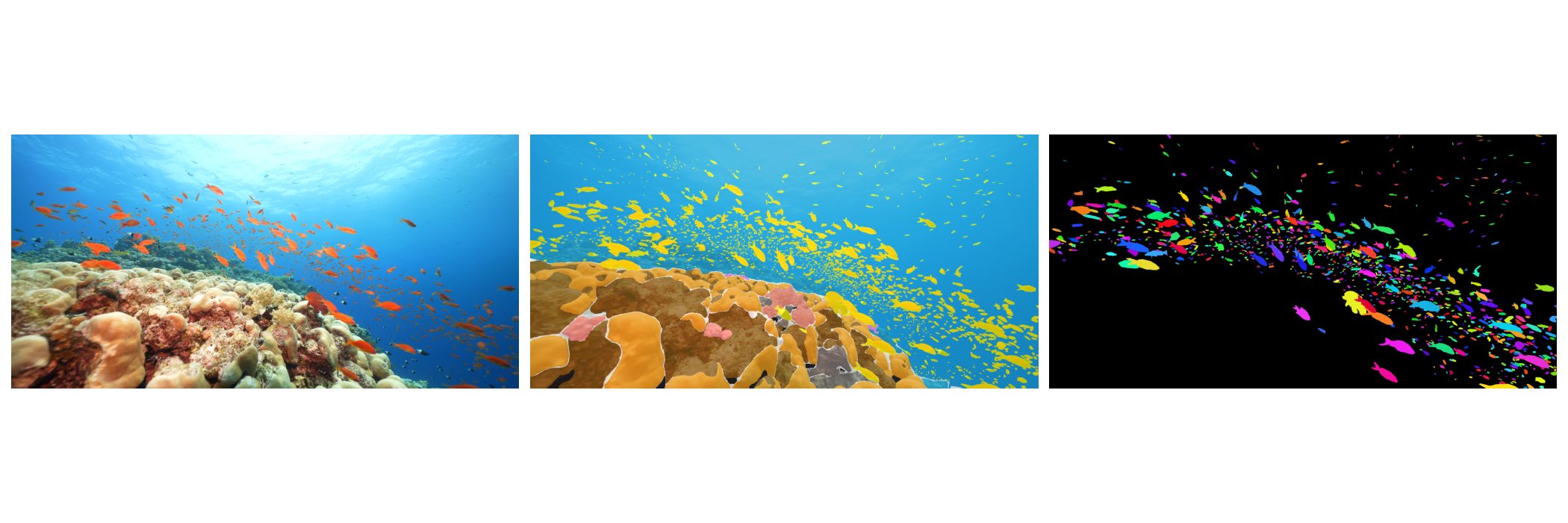}
\tiny
  \makebox[0.14\linewidth][c]{More Scenes}
  \hspace{1.0cm}
  \makebox[0.33\linewidth][c]{Improved Fine-Grained Segmentation}
  \hspace{1.0cm}
  \makebox[0.14\linewidth][c]{Exhaustive Fish Instances}
\end{center}
\begin{abstract}
In order to design conservation and restoration strategies to counter the global decline of coral reefs, ecological monitoring of reefs needs to be scaled up dramatically. Computer vision methods are increasingly used to tackle the vast amount of data: as the paradigm of data collection in reefs shifts from highly standardized and constrained survey images to unconstrained imagery on scalable platforms, it is necessary to design machine learning methods that help to get a fine-grained understanding of reefs from general-purpose reef imagery. This paper provides CoralscapesV2, an extension of the Coralscapes dataset for general-purpose visual scene understanding in reefs. CoralscapesV2 increases the dataset size, scope, label completeness and quality for semantic segmentation, and extends the number of classes from 39 to 95 fine-grained visual categories. Furthermore, CoralscapesV2 provides 65k \textbf{exhaustive} fish instance mask annotations, meticulously annotated to completeness by using the video, revealing that annotation of fish based on only static images is insufficient. CoralscapesV2 is the first dataset for panoptic segmentation in coral reefs, capturing a wide range of scenarios in the wild, posing a challenging benchmark for contemporary semantic segmentation and instance segmentation models. CoralscapesV2 is an important step towards general-purpose panoptic segmentation in coral reefs, which has substantial implications for scaling up coral reef monitoring, as it can be employed in a wide range of applications from benthic cover mapping from robot or handheld videos to designing methods for automated quantification and understanding of fish behavior and fish-reef interactions.
\end{abstract}

\section{Introduction}


Coral reefs are under unprecedented threat from anthropogenic climate change and local stressors such as overfishing and pollution. Recurrent mass-bleaching events driven by marine heatwaves have caused widespread coral mortality over the past decades \cite{hughes2017bleaching}, restructuring reef assemblages \cite{hughes2018transforms} and eroding the capacity of reefs to provide the ecosystem services on which hundreds of millions of people depend \cite{eddy2021decline}. Yet the dynamics of reef decline and recovery remain poorly understood and vary substantially across regions \cite{hughes2018transforms}, while only a small fraction of the world's reefs are actively and regularly monitored \cite{gcrmnreport}. Designing effective conservation and restoration strategies therefore requires ecological monitoring that is simultaneously scaled up in spatial and temporal coverage and deepened in the fine-grained ecological detail it captures.

Computer vision is increasingly used in reef monitoring to process the vast amounts of imagery collected. As AI-based vision tools mature, the field is shifting away from highly standardized acquisition protocols (\eg photo quadrats and orthophotos) towards \textbf{general-purpose} analysis of reef imagery that can be applied flexibly across survey types and data-collection platforms: diver-held cameras \cite{sauder2024scalable, sauder2025rapid}, ROV/AUV platforms \cite{girdhar2023curee, raine2025ai}, and static underwater cameras \cite{lilkendey2024herbivorous,khiem2025novel}.

Most existing datasets target isolated use cases, leaving a real need for high-quality, general-purpose datasets. Coralscapes \cite{sauder2025coralscapes} was the first dataset explicitly designed to benchmark general-purpose segmentation in reefs, providing densely annotated imagery from 35 dive sites across five countries in the Red Sea region. However, Coralscapes emphasizes the benthos, treating fish merely as a single semantic class to be masked out of benthic mapping.

A key component of fine-grained reef understanding, however, is the study of the fish populations reefs host and their entangled dynamics. A variety of datasets and methods for detecting, tracking, and taxonomically identifying fish in images and video have been proposed \cite{joly2015lifeclef,boom2014fish4knowledge, kavasidis2013f4k,cai2025corefish}. However, these are either limited to static images, incompletely annotated, from outside the coral-reef domain, or lacking instance masks, limiting their usefulness for real-world ecological applications such as fish counting or biomass estimation on reefs.

This paper proposes CoralscapesV2, an extension of Coralscapes that bridges this gap and improves the original dataset along every measurable axis:
\begin{enumerate}
    \item \textbf{Increased Size, Completeness, and Quality} We increase the number of annotated images (2075 to 2433), the number of annotated polygons (174k to 270k), and the annotation density (82.9\% to 86.8\%), while broadening the diversity of scenes across more reef sites in the Red Sea region (35 to 45). Annotation masks of existing images are further refined in terms of density and quality.
    \item \textbf{Fine-Grained Visual Categories} A refined label set of 95 fine-grained classes captures more taxa, coral morphologies along with their health status, as well as new auxiliary classes.
    \item \textbf{Exhaustive Fish Instance Annotations}
    65k fish are exhaustively annotated on each frame using the video.
\end{enumerate}

Together, these improvements make CoralscapesV2 the first panoptic segmentation dataset for coral reefs. This unlocks applications that were previously out of reach for reef imagery, from joint benthic-cover mapping and fish quantification on unconstrained diver or robot video to the study of fish behavior and fish-reef interactions. The remainder of this paper reviews related datasets (Sec.~\ref{sec:related}), details CoralscapesV2 and its annotation effort (Sec.~\ref{sec:dataset}), and benchmarks contemporary semantic- and instance-segmentation models (Sec.~\ref{sec:experiments}).
\AddToShipoutPictureFG*{%
  \AtPageLowerLeft{%
    \raisebox{15mm}{%
      \makebox[\paperwidth][c]{%
        Data and models available through: \href{https://josauder.github.io/coralscapesv2/}{\scriptsize https://josauder.github.io/coralscapesv2/}%
      }%
    }%
  }%
}
\section{Related Work}
\label{sec:related}
\vspace{-4pt}

\subsection{Coral Reef Image Datasets}
    \vspace{-2pt}

Annotation of coral reefs at scale is challenging: many corals show high morphological plasticity and taxa can not be identified to species or genus level from imagery. Some genera or species can only be identified with high-resolution close-up images of the corallites, and some simply can not be visually identified. Annotating corals across image quality and water conditions, falling back to the growth form when taxonomy is not clear, and disambiguating other benthic classes (e.g. sponges) requires substantial domain expertise. As such, large-scale annotation of reefs is limitingly expensive, and the landscape of datasets is scattered around conventions, tools, and applications.

Most coral reef scientists and conservation practitioners interact with imagery through platforms such as CoralNet \cite{beijbom2012moorea}, MERMAID \cite{datamermaid}, or ReefCloud \cite{reefcloud}, uploading images and annotating them with sparse point labels that are used to train classification models. Consequently, most annotated reef data exists as sparse point labels, as in the Moorea Labeled Corals \cite{beijbom2012moorea, beijbom2015towards} and Catlin Seaview \cite{gonzalez2019seaview} datasets. A smaller set of datasets provides dense masks rather than sparse points, such as UCSD mosaics \cite{edwards2017large} or Benthos \cite{yuval2021repeatable}, but these are confined to the orthomosaic domain and limited in scale; CoralSCOP \cite{zheng2024coralscop} reaches large scale but only distinguishes coral from non-coral, on noisy imagery and labels.

Coralscapes~\cite{sauder2025coralscapes} proposed the first general-purpose dense segmentation dataset, with 2075 densely annotated video frames across 39 classes from 35 reef sites in the Red Sea region. It follows the structure of the Cityscapes dataset~\cite{cordts2016cityscapes}: each $1024\times2048$px image is a frame from a video, with the preceding 20 and following 10 frames available. However, Coralscapes frames were annotated statically, with no temporal context and fish captured only as a single semantic class, instead of as individual instances. CoralscapesV2 directly extends this line of work, refining the label set to 95 classes and, crucially, adding exhaustive fish \emph{instance} masks annotated with video context.

\subsection{Fish Datasets}
\vspace{-4pt}
\begin{table*}[t!]
    \centering
    \scriptsize
    \setlength{\tabcolsep}{4pt}
    \caption{Fish datasets, grouped by annotation type and sorted by size within each group. \emph{Camera}: image-only vs.\ video (static/moving/mixed). \emph{Exhaustive}: whether all fish in a labeled frame are annotated. $\dagger$~= total over all categories (not fish-only). 
    }
    \label{tab:fishdatasets}
    \vspace{-4pt}
    \resizebox{\textwidth}{!}{%
    \begin{tabular}{llllrrc}
        \toprule
        Dataset & Ecosystem & Camera & Annotation Type & Images/frames & Annotations & Exhaustive \\
        \midrule
        \multicolumn{7}{l}{\textit{Bounding boxes}} \\
        FishTrack23~\cite{dawkins2024fishtrack23} & Mixed & \textbf{Video (mixed)} & Box + track & -- & $\sim$850{,}000 & partial \\
        Orange Chromide~\cite{vijayalakshmi2024orangechromide} & Freshwater & Images & Box & 586 & 10{,}607 & \textcolor{red}{N} \\
        Roboflow Fish~\cite{roboflow_fish} & Mixed & Images & Box & 680 & 3{,}142 & \textcolor{red}{N} \\
        Labeled Fishes~\cite{cutter2015labeledfishes} & Deep sea & \textbf{Video (moving)} & Box & 929 & 1{,}005 & \textcolor{red}{N} \\
        TORSI~\cite{scoulding2025torsi} & Deep sea & Images & Box & 1{,}051 & 14{,}414 & \textcolor{red}{N} \\
        Project Natick~\cite{natick2018} & Marine & \textbf{Video (static)} & Box & 1{,}118 & 998 & \textcolor{red}{N} \\
        OzFish~\cite{marrable2022ozfish} & Mixed & \textbf{Video (static)} & Box & $\sim$1{,}800 & $\sim$45{,}000 & \textcolor{blue}{Y} \\
        Deep Vision~\cite{allken2021deepvision} & Pelagic & \textbf{Video (moving)} & Box & 1{,}875 & 4{,}834 & \textcolor{red}{N} \\
        AAU Zebrafish~\cite{haurum2020zebrafish} & Aquaculture & \textbf{Video (static)} & Box + track & 2{,}224 & $\sim$6{,}672 & \textcolor{blue}{Y} \\
        J-EDI JODD~\cite{nishio2026jodd} & Deep sea & \textbf{Video (moving)} & Box & 8{,}151 & 15{,}621$\dagger$ & \textcolor{red}{N} \\
        Brackish~\cite{pedersen2019brackish} & Brackish & \textbf{Video (static)} & Box & 14{,}518 & 12{,}797 & \textcolor{blue}{Y} \\
        CoReFish~\cite{cai2025corefish} & Coral reef & \textbf{Video (moving)} & Box + track & 14{,}580 & 98{,}267 & partial \\
        FishCLEF-2015~\cite{joly2015lifeclef} & Coral reef & \textbf{Video (static)} & Box & $>$20{,}000 & $>$14{,}000 & \textcolor{red}{N} \\
        Marine Detect~\cite{marinedetect} & Mixed & Images & Box & $\sim$21{,}125 & -- & \textcolor{red}{N} \\
        N-MARINE~\cite{ayyagari2025nmarine} & Deep sea & \textbf{Video (static)} & Box & 23{,}936 & -- & \textcolor{blue}{Y} \\
        OBSEA~\cite{francescangeli2023obsea} & Coastal & Images & Box & 33{,}805 & 69{,}917 & \textcolor{red}{N} \\
        Kakadu~\cite{jansen2024kakadu} & Freshwater & \textbf{Video (static)} & Box & 44{,}112 & 82{,}904 & \textcolor{red}{N} \\
        MFT25~\cite{li2026mft25} & Aquaculture & \textbf{Video (static)} & Box + track & 48{,}066 & 408{,}578 & \textcolor{blue}{Y} \\
        River Herring~\cite{mitseagrant_riverherring} & Freshwater & \textbf{Video (static)} & Box & $\sim$60{,}000 & 91{,}482 & \textcolor{blue}{Y} \\
        Puget Sound~\cite{farrell2023pugetsound} & Estuary & \textbf{Video (static)} & Box & 77{,}739 & 67{,}990$\dagger$ & \textcolor{blue}{Y} \\
        FathomNet~\cite{katija2022fathomnet} & Deep sea & Images & Box & 84{,}454 & 175{,}873$\dagger$ & \textcolor{red}{N} \\
        FishNet~\cite{khan2023fishnet} & Mixed & Images & Box & 94{,}532 & 114{,}375 & \textcolor{red}{N} \\
        Salmon CV~\cite{atlas2023salmon} & Freshwater & \textbf{Video (static)} & Box + track & 532{,}000 & $\sim$503{,}000 & \textcolor{blue}{Y} \\
        \midrule
        \multicolumn{7}{l}{\textit{Semantic masks (no instance)}} \\
        DeepFish~\cite{saleh2020deepfish} & Mixed & \textbf{Video (static)} & Semantic & 620 & -- & \textcolor{red}{N} \\
        SUIM~\cite{islam2020suim} & Mixed & Images & Semantic & 1{,}635 & -- & \textcolor{red}{N} \\
        Coralscapes~\cite{sauder2025coralscapes} & Coral reef & \textbf{Video (moving)} & Semantic & 2{,}075 & $\sim$22{,}000 fish & \textcolor{red}{N} \\
        \midrule
        \multicolumn{7}{l}{\textit{Instance masks}} \\
        Fish4Knowledge~\cite{kavasidis2013f4k} & Coral reef & \textbf{Video (static)} & \textbf{Instance} + point track & 17 videos & -- & \textcolor{red}{N} \\
        PomerFish~\cite{shi2026pomerfish} & Freshwater & \textbf{Video (mixed)} &\textbf{Instance} & 1{,}115 & 1{,}038 & \textcolor{red}{N} \\
        DeepFish-Inst~\cite{garciadurso2022deepfishinstance} & Aquaculture & Images & \textbf{Instance} & 1{,}291 & 7{,}339 & \textcolor{red}{N} \\
        WaterMask/UIIS~\cite{lian2023watermask} & Mixed & Images & \textbf{Instance} & 4{,}628 & -- & \textcolor{red}{N} \\
        TrashCan~\cite{hong2020trashcan} & Deep sea & \textbf{Video (moving)} & \textbf{Instance} & 7{,}212 & -- & \textcolor{red}{N} \\
        UIIS10K~\cite{li2025uiis10k} & Mixed & Images & \textbf{Instance} & 10{,}048 & 41{,}862$\dagger$ & \textcolor{red}{N} \\
        Fish-Occlusion~\cite{zhang2026fishocclusion} & Aquaculture & Images & \textbf{Instance} & 14{,}376 & 144{,}894 & \textcolor{red}{N} \\
        USIS16K~\cite{hong2025usis16k} & Mixed & Images & \textbf{Instance} & 16{,}151 & -- & \textcolor{red}{N} \\
        \midrule
        \textbf{CoralscapesV2} & Coral reef & \textbf{Video (moving)} & \textbf{Instance} & \textbf{2433} & \textbf{65{,}738 fish} & \textbf{\textcolor{blue}{Y}} \\
        \bottomrule
    \end{tabular}%
    }
    \vspace{-10pt}

\end{table*}

Beyond the benthos, a large and growing body of datasets targets fish. We organize this landscape (Table~\ref{tab:fishdatasets}) along the three central axes: whether annotations are per-fish \emph{instance masks}, whether every fish in a frame is \emph{exhaustively} labeled, and whether the data covers \emph{coral-reef} scenes with \emph{video} context.

Annotating fish exhaustively is challenging at scale. Labeling only the obvious, salient fish is comparatively easy, but many important applications -- such as fish tracking for abundance estimation or behavior analysis -- require that \emph{every} visible fish in an image be recognized. Exhaustive annotation is costly because it must be performed with video rather than single frames, which is slow because many fish appear blurred, occluded, or as tiny objects, and can be hard to distinguish from other suspended matter such as marine snow or under poor visibility. As a result, most fish datasets provide only image-level labels or bounding boxes \cite{zhuang2018wildfish, khan2023fishnet}; several of the largest are exhaustively annotated but remain box-only and outside reefs. Per-instance masks are far rarer and, where they exist, come from freshwater, aquaculture, or domain-generic underwater settings and are non-exhaustive or restricted to salient objects \cite{shi2026pomerfish, garciadurso2022deepfishinstance, zhang2026fishocclusion, lian2023watermask, li2025uiis10k, hong2025usis16k}. Within coral reefs, existing fish datasets stop short of exhaustive per-fish instance masks: FishCLEF-2015~\cite{joly2015lifeclef} and Fish4Knowledge~\cite{boom2014fish4knowledge, kavasidis2013f4k} provide only bounding boxes or instance masks from static-camera video, and the recent CoReFish~\cite{cai2025corefish} -- closest to ours in capturing diver ego-motion reef video -- provides bounding boxes and tracks rather than instance masks. As the table shows, no existing dataset combines coral-reef scenes, exhaustive per-fish instance masks, and video context; CoralscapesV2 is, to our knowledge, the first to provide all three.

\section{Dataset}
\label{sec:dataset}
\begin{figure}[b!]
    \vspace{-5pt}
    \centering
    \includegraphics[width=0.495\linewidth,trim={10px 0px 3px 18px},clip]{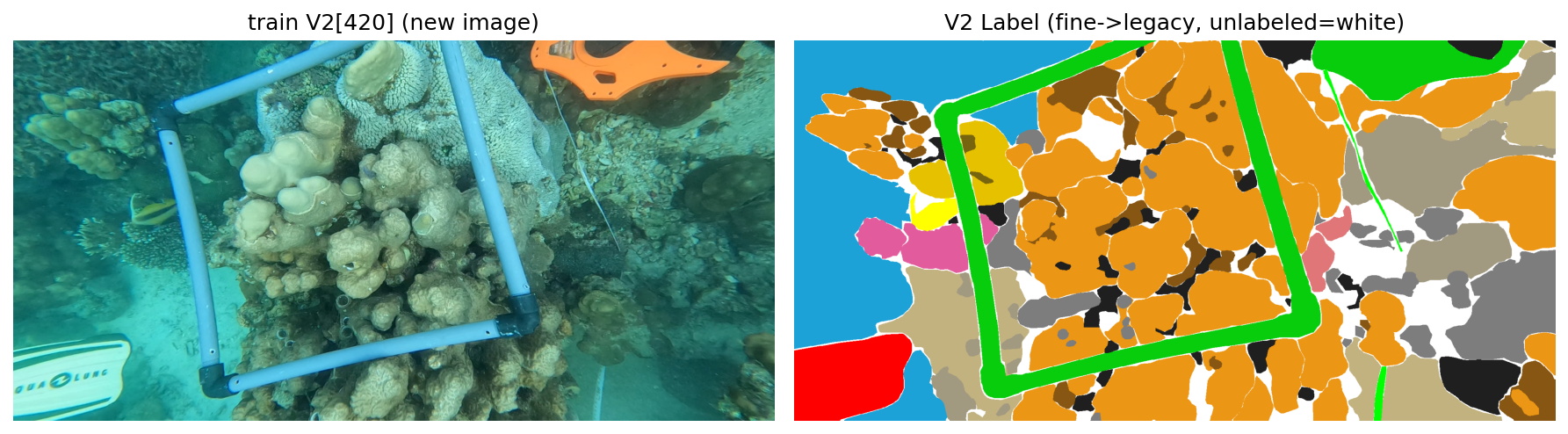}
    \includegraphics[width=0.495\linewidth,trim={3px 0px 10px 18px},clip]{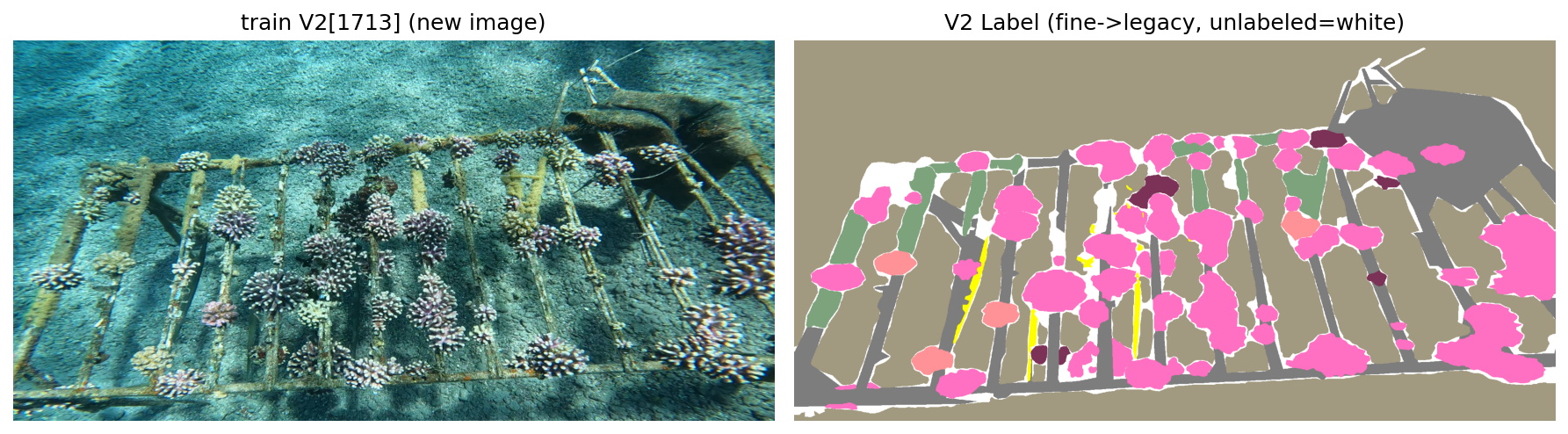}
    \includegraphics[width=0.495\linewidth,trim={10px 0px 3px 18px},clip]{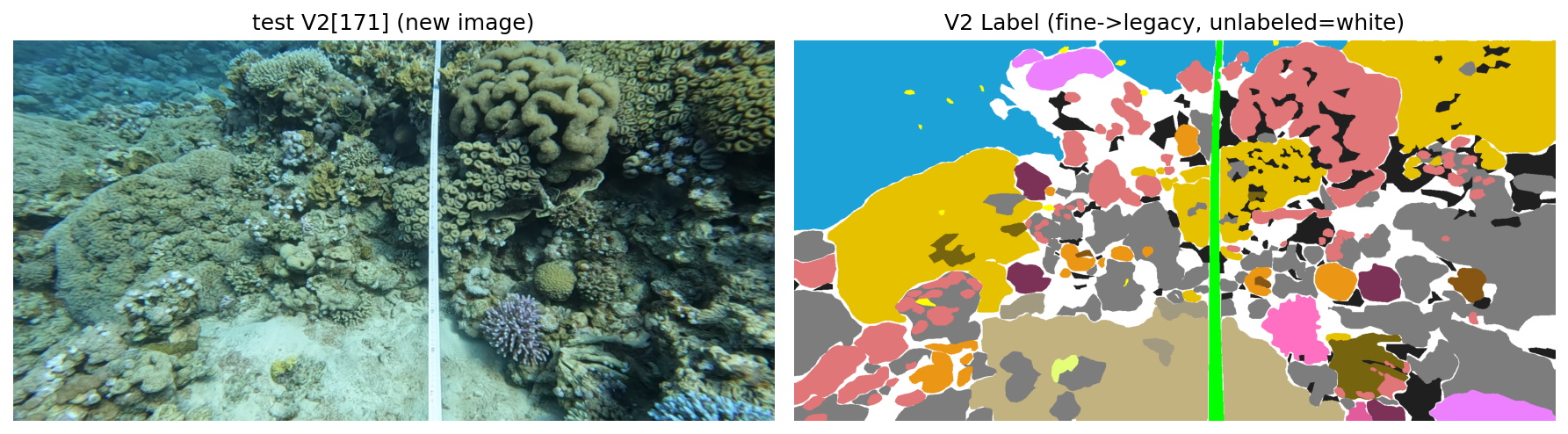}
    \includegraphics[width=0.495\linewidth,trim={3px 0px 10px 18px},clip]{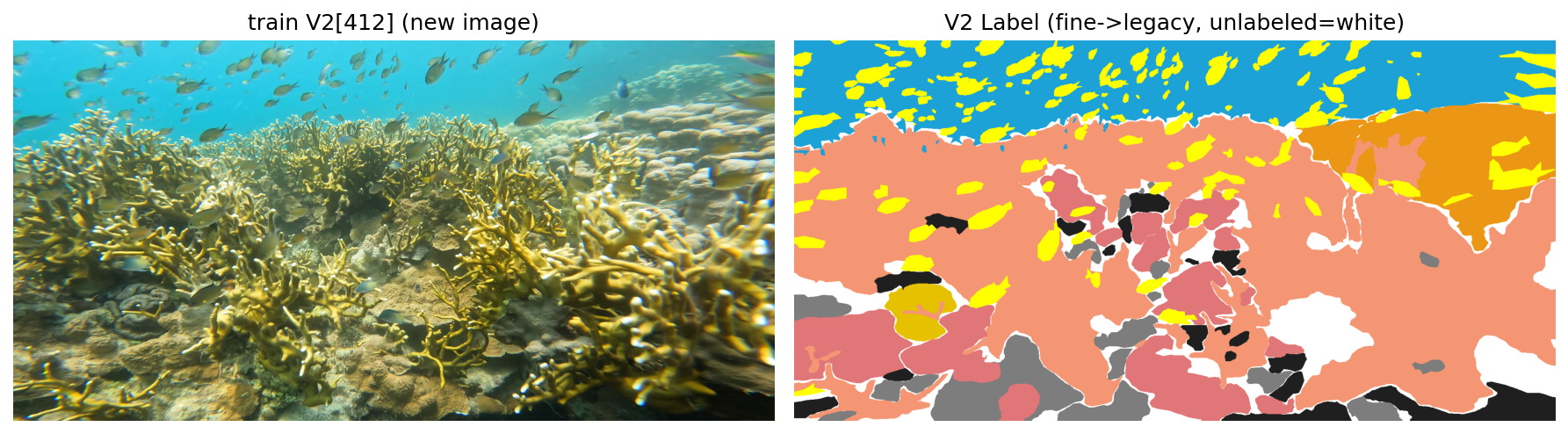}
    \includegraphics[width=0.495\linewidth,trim={10px 0px 3px 18px},clip]{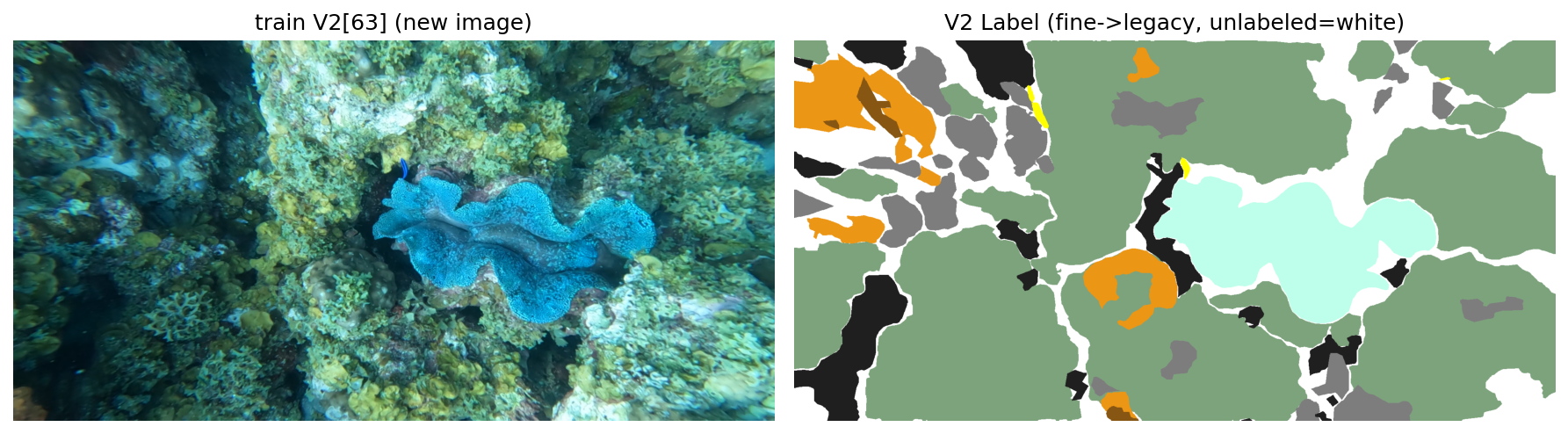}
    \includegraphics[width=0.495\linewidth,trim={3px 0px 10px 18px},clip]{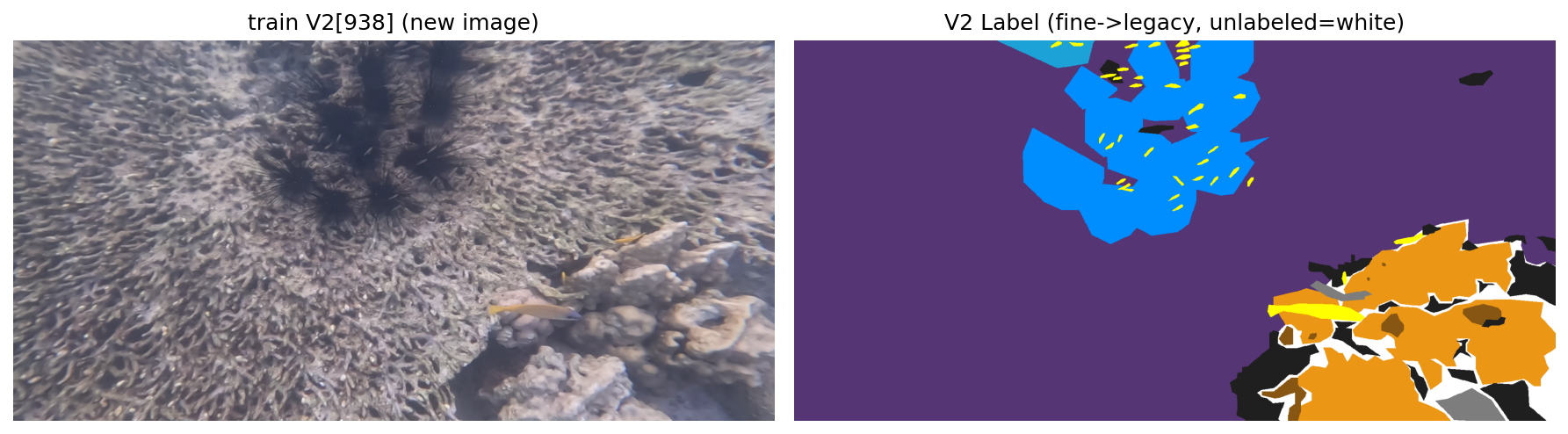}
    \includegraphics[width=0.495\linewidth,trim={10px 0px 3px 18px},clip]{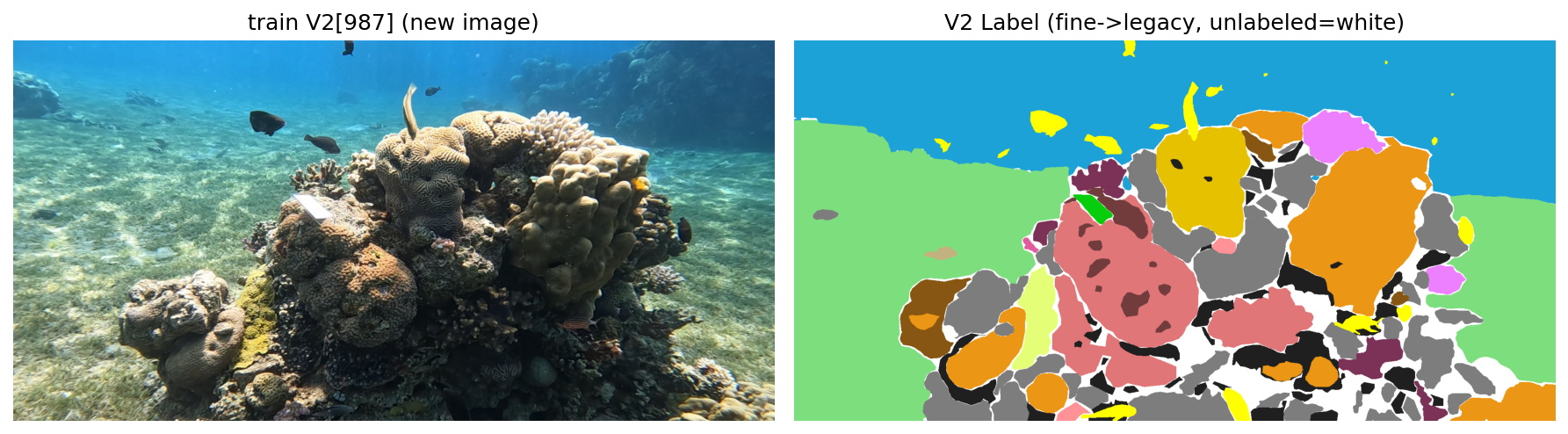}
    \includegraphics[width=0.495\linewidth,trim={3px 0px 10px 18px},clip]{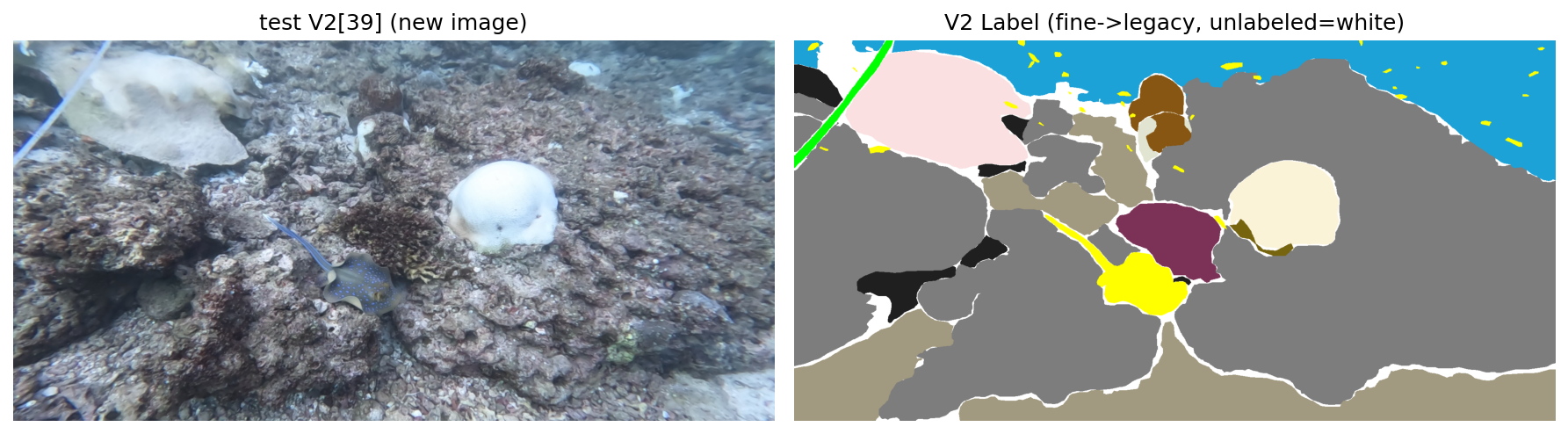}
    \includegraphics[width=0.495\linewidth,trim={10px 0px 3px 18px},clip]{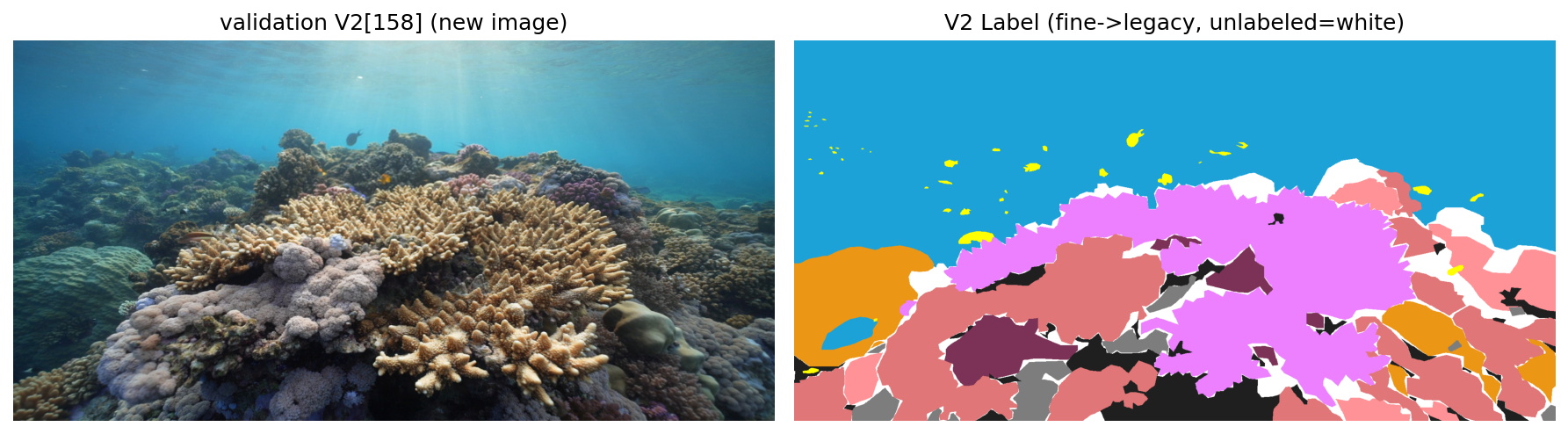}
    \includegraphics[width=0.495\linewidth,trim={3px 0px 10px 18px},clip]{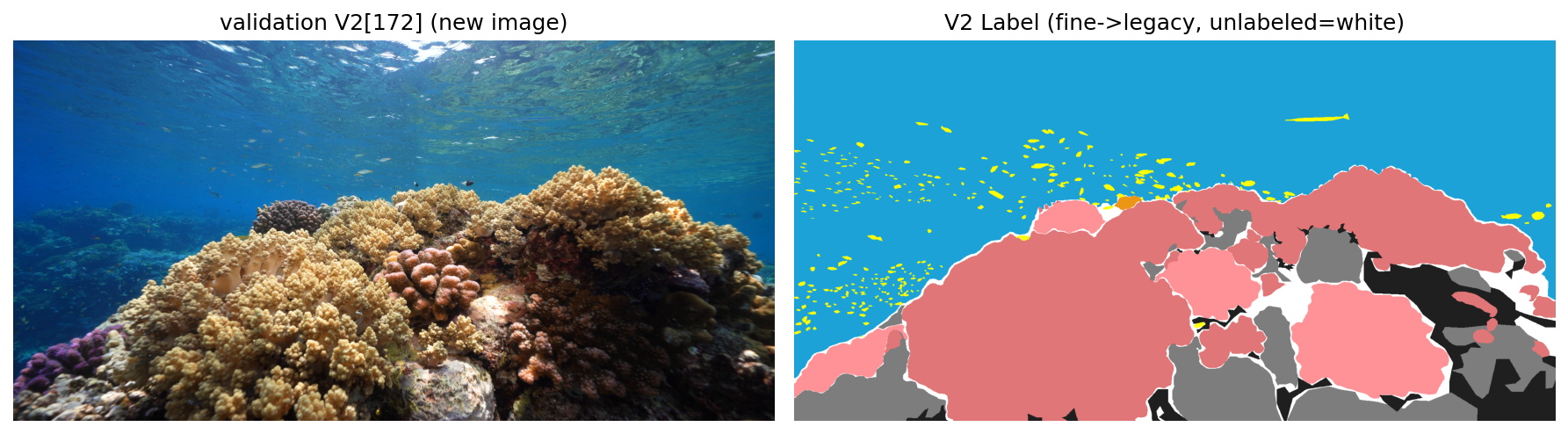}
    \includegraphics[width=\linewidth]{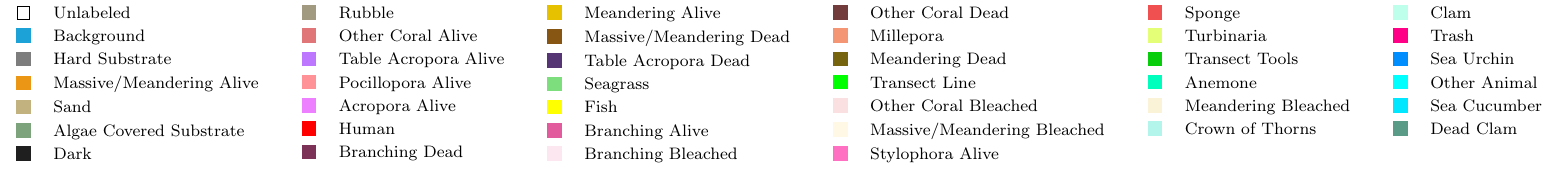}
       \vspace{-5pt}
    \caption{
    CoralscapesV2 adds 358 new annotated video frames, with an emphasis on increasing Coralscapes' diversity of scenes, for example by including photo quadrats and coral restoration infrastructure (top row), or images from a high-quality camera (bottom row).}
        \vspace{-5pt}

    \label{fig:new_images}
\end{figure}

CoralscapesV2 improves upon the original Coralscapes along every measurable axis. In principle, the main improvements are \textbf{Increased Size, Completeness, and Quality} of the dataset, \textbf{Fine-Grained Visual Categories} and \textbf{Exhaustive Fish Instance Annotations}. The remainder of this Section details these improvements and the involved annotation effort.

\begin{figure}
    \centering
    \includegraphics[width=0.49\linewidth]{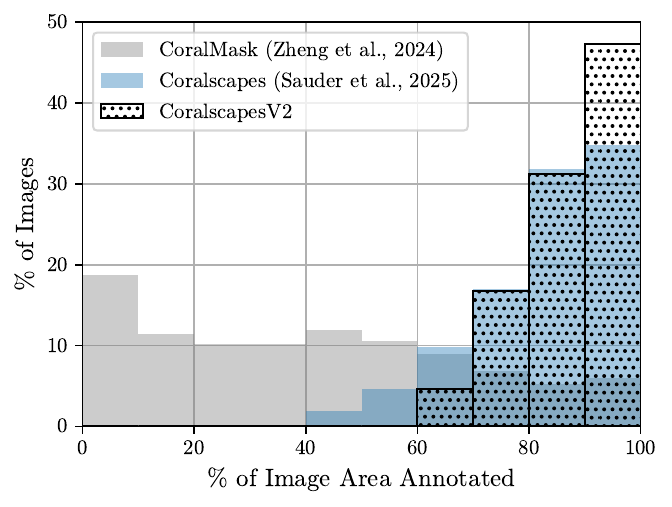}
    \includegraphics[width=0.49\linewidth]{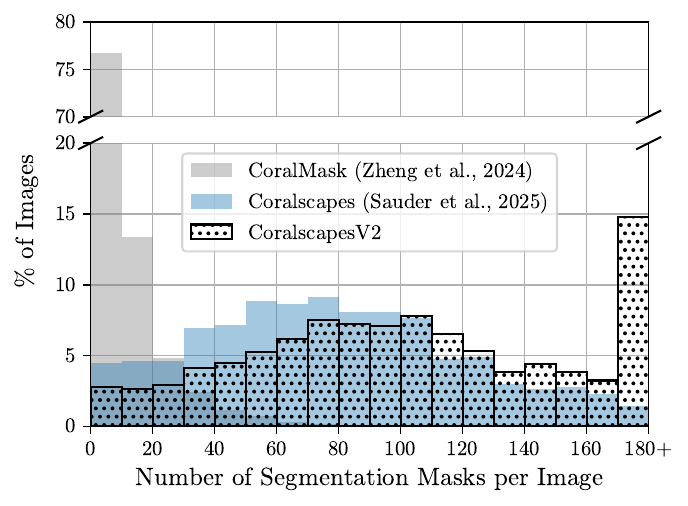}
    \vspace{-5pt}
    \caption{CoralscapesV2 increases the annotation density and complexity.}
    \vspace{-5pt}
    \label{fig:histograms}
\end{figure}

\subsection{Increased Size, Completeness, and Quality}
\vspace{-3pt}

CoralscapesV2 increases the size of the Coralscapes dataset along every dimension. The total number of annotated video frames is increased from 2075 to 2433 images (Fig.~\ref{fig:new_images}). All existing and new imagery was taken during scientific expeditions and monitoring workshops in reefs of the Red Sea region (Jordan, Djibouti, Eritrea, Sudan, and Israel) between August 2021 and October 2025. Of the 358 new frames, 246 are from reef sites that were present in Coralscapes V1, and the remaining frames are from 10 new reef sites, which are split into train/validation/test splits of 1790/170/473 frames along geographically disjoint reef sites\footnote{As in the original Coralscapes dataset, the geolocations of individual sites are anonymized in accordance with the sovereignty of local monitoring efforts and to protect reefs from targeted tourism or overfishing pressure.}. The splits of the reef sites match those of the original Coralscapes dataset (i.e. any image that existed in V1 will be in the same split in V2). Of the new annotated frames, 303 are taken with GoPro Hero 10 cameras (as all of Coralscapes V1), and the remainder with a Sony 7r4 camera (12-24mm f4 G lens), which produces higher-quality footage.

\begin{figure}[t!]    \vspace{-5pt}
\centering
\begin{tabular}{>{\centering\arraybackslash}m{0.04\textwidth} m{0.94\textwidth}}
(a) &
\includegraphics[width=\linewidth]{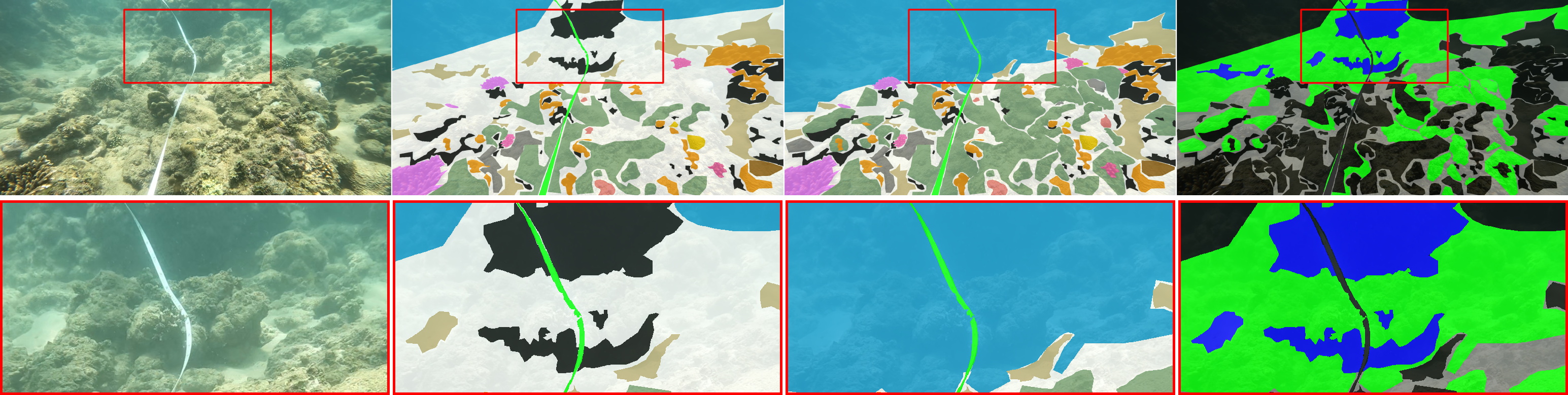} \\[10ex]

(b) &
\includegraphics[width=\linewidth]{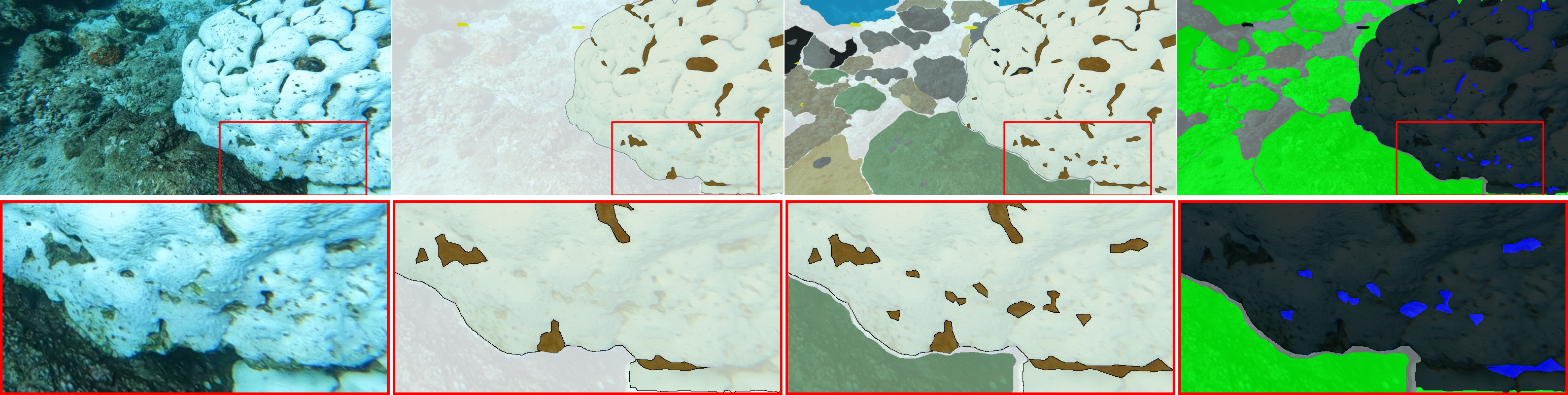} \\[10ex]

(c) &
\includegraphics[width=\linewidth]{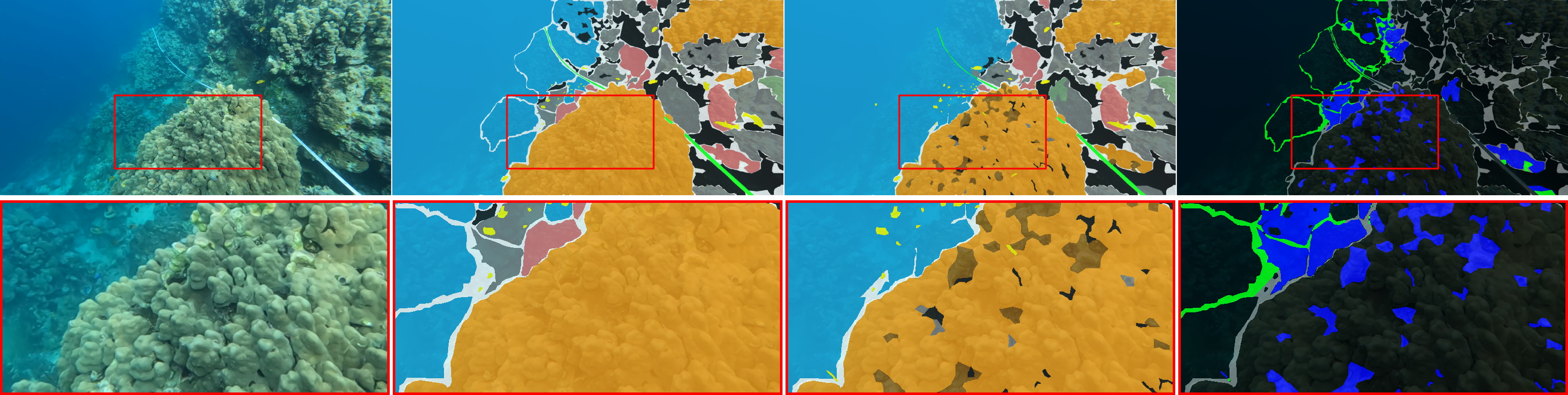}
\end{tabular}
\raggedleft
\includegraphics[width=0.94\linewidth, trim={6px 20px 10px 20px},clip]{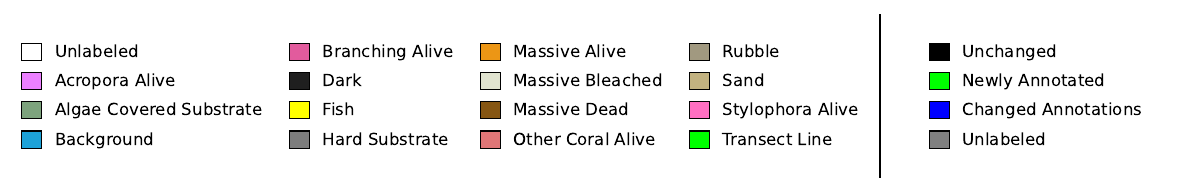}
\caption{CoralscapesV2 improves the segmentation targets of existing frames: common improvements are annotating substantially more of the frame area (a, b), applying auxiliary classes like `background' and `dark' more consistently (a), alleviating image-border artifacts from legacy annotations with SAM (b), and filling in small polygons on over-annotated images, including dead and dark patches (b, c). For better visualization, the massive bleached coral mask is outlined in (b).}
\vspace{-10pt}
\label{fig:improved_images}
\end{figure}

Furthermore, CoralscapesV2 increases the annotation \textbf{density}, as shown by Figure~\ref{fig:histograms}, across both existing and newly added images. The average area of images that is annotated is increased from 82.9\% to 86.8\%, and the minimum annotated area for each image from 40\% to 60\%. In total, CoralscapesV2 increases the total number of user-annotated segmentation masks from 174k to 270k. Besides increasing the number of annotated pixels, every frame's annotation mask was thoroughly reviewed and potentially improved in terms of \textbf{quality}, as showcased in Figure~\ref{fig:improved_images}.
In particular, the annotation of the original Coralscapes dataset was performed with substantial use of Segment Anything Model (SAM) \cite{kirillov2023sam}, which led to border artifacts, and encouraged over-segmentation: often, the masks for coral colonies included dead coral patches or dark regions, which should be appropriately annotated into their respective classes. Furthermore, the auxiliary `background' and `dark' classes, which are innately subjective but crucial for forcing a non-prediction of benthic classes, are applied more consistently in CoralscapesV2. Beyond that, isolated wrong annotations were corrected and all visible instances of liagora algae (often appearing distinctly white in Coralscapes) were moved from the wrong `other animal' class to the `algae covered substrate' class.

\subsection{Fine-Grained Visual Categories}
\vspace{-3pt}

The original Coralscapes dataset summarized all annotated classes into 39 label classes that were selected such that classes are well-represented in the train, validation and test splits. CoralscapesV2 relaxes this requirement, and provides 95 fine-grained visual categories that are consistently annotated. The remainder of this Section describes the newly added classes.\\

\noindent\textbf{More Taxa \& Morphotypes} \\
\noindent
CoralscapesV2 separates clearly identifiable coral taxa that were previously summarized. In particular, CoralscapesV2 has classes for visually clearly identifiable taxa, spanning both hard corals (`lobophyllidae', `porites', `fungiidae', `pavona', `galaxea', `goniopora', `seriatopora'), and soft coral taxa (`alcyoniidae',`nephtheidae', `xeniidae'). For classes with a clearly distinct morphology but where the taxon can not be identified clearly, new morphologic classes for `brain coral' and `thin plate/encrusting coral' are introduced. Two catch-all classes, `hard coral alive' and `soft coral alive', absorb any coral not assigned a finer label. This guarantees full coverage while preserving the ecologically important hard–soft coral distinction.
Furthermore, classes for the visually distinct black coral `cirrhipathes' and sabellid `feather worm' are included. 

\noindent The annotation strategy prioritizes correctness over taxonomic depth: labels are given in a non-speculative manner. 
For example, species in the `pavona' genus can take many growth forms, some of which can be hard to identify as pavona without a high-resolution view of the corallites. In these ambiguous cases, the morphologic label is applied. However, other species (e.g. `pavona cactus' or `pavona decussata') can be confidently identified within the genus pavona, but not distinguished from each other on CoralscapesV2 imagery.\\

\noindent\textbf{Fine-Grained Coral Composite Health Classes}\\
\noindent On top of providing additional taxa, CoralscapesV2 also provides composite health classes (bleached, dead, algae covered) for taxonomic classes.

\begin{figure}[t!]
    \centering
    \includegraphics[width=0.495\linewidth,trim={0px 0px 0px 23px},clip]{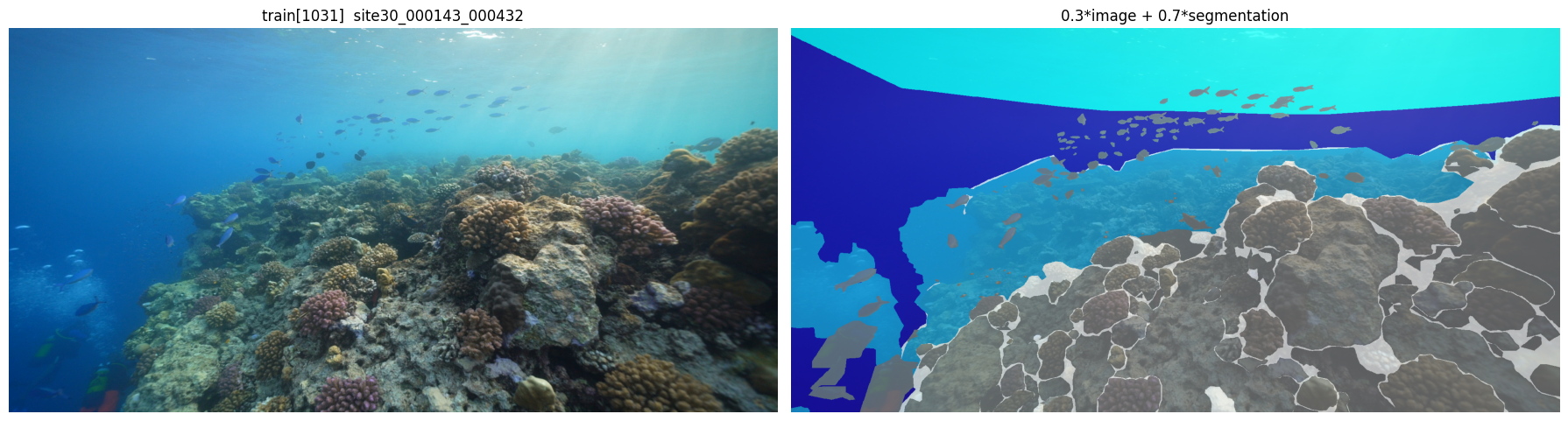}
    \includegraphics[width=0.495\linewidth,trim={0px 0px 0px 23px},clip]{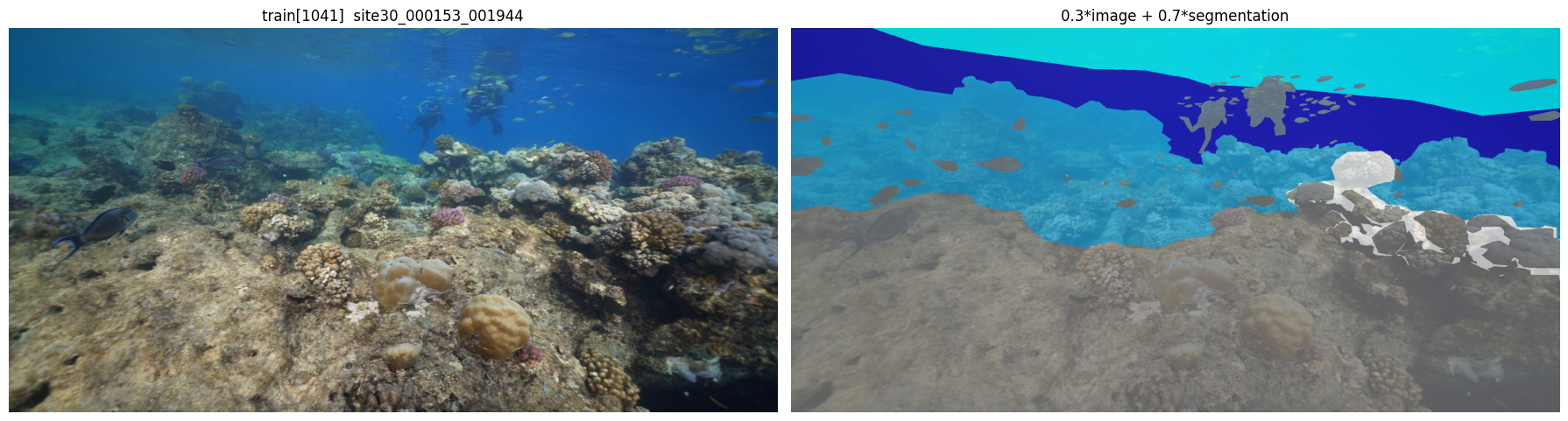}
    \includegraphics[width=0.495\linewidth,trim={0px 0px 0px 23px},clip]{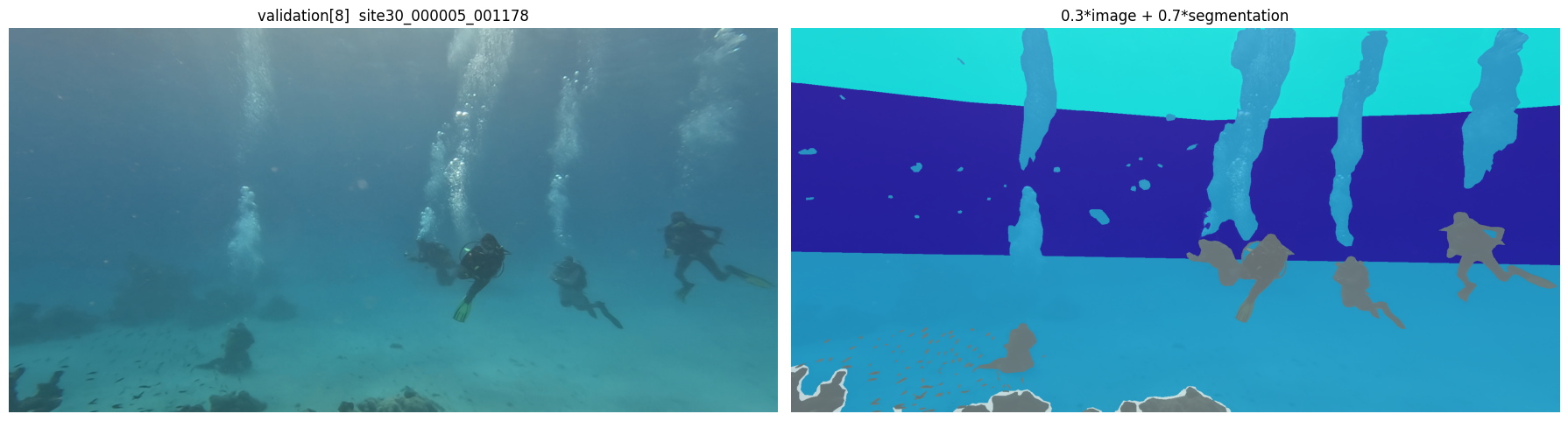}
    \includegraphics[width=0.495\linewidth,trim={0px 0px 0px 23px},clip]{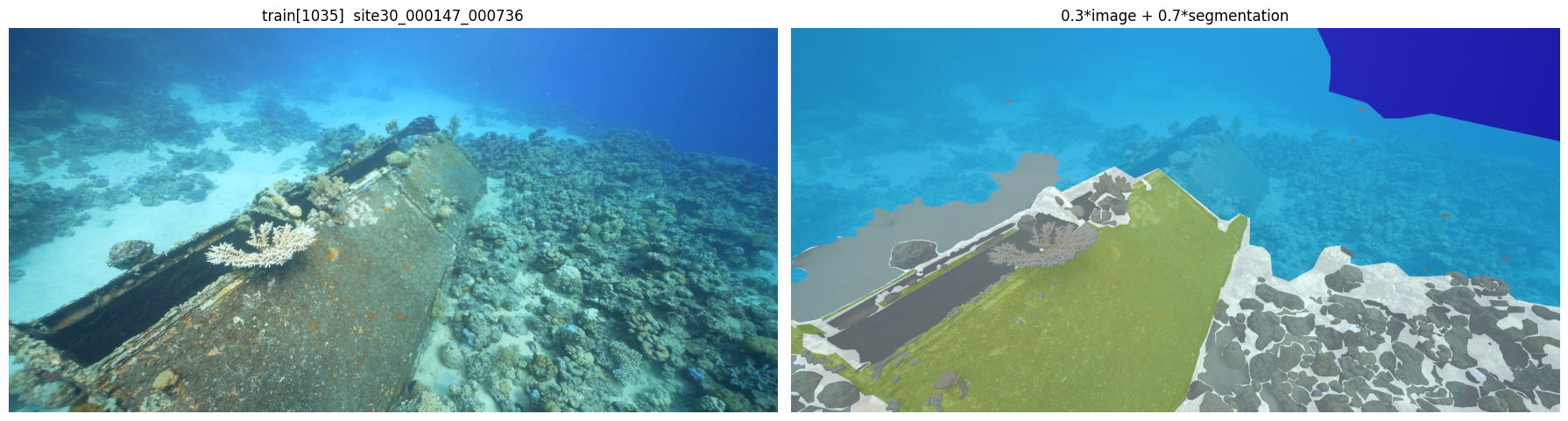}
    \includegraphics[width=0.495\linewidth,trim={0px 0px 0px 23px},clip]{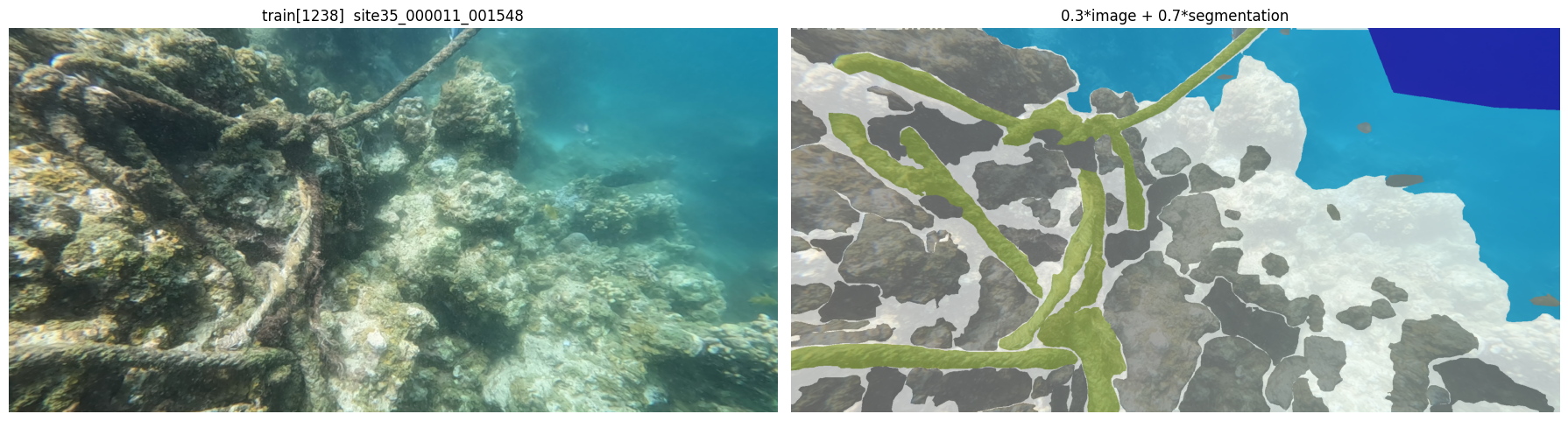}
    \includegraphics[width=0.495\linewidth,trim={0px 0px 0px 23px},clip]{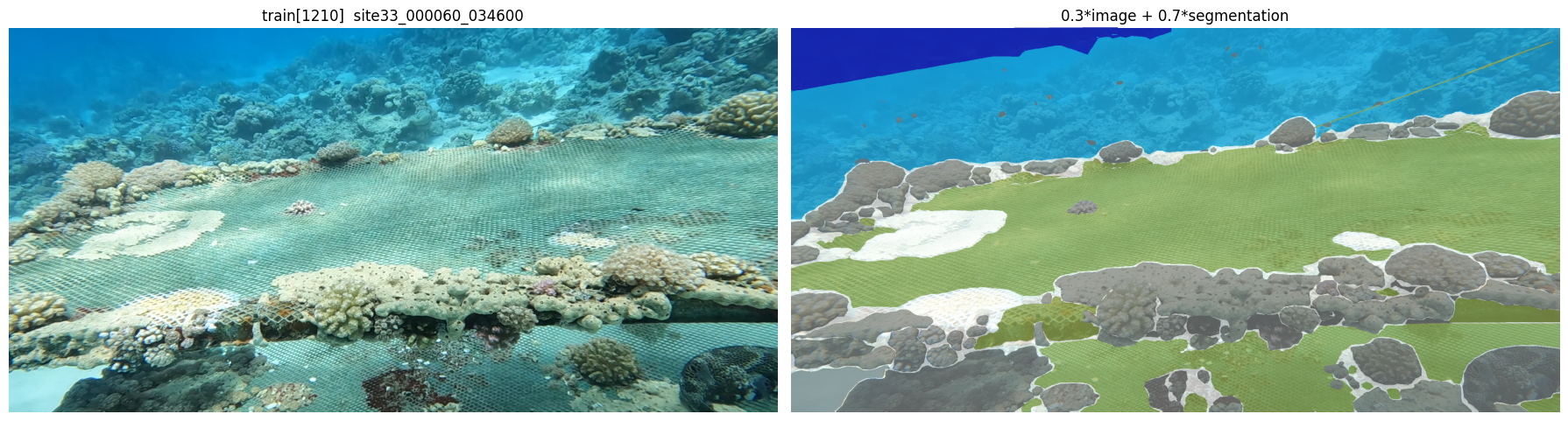}
    \includegraphics[width=0.65\linewidth,trim={0px 18px 0px 15px},clip]{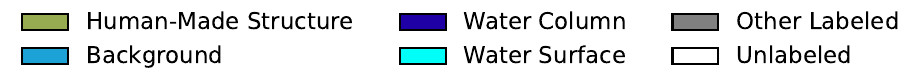}
    \caption{Samples emphasizing the disambiguated semantics of the new auxiliary classes: `water column' designates image areas with an unobstructed view of the water (no substrate, fish, divers, or marine snow visible), `water surface' denotes shimmering regions from surface waves, and `background' catches all other image areas where blur/visibility prevent a speculation-free assessment of coral taxon or health status.}
    \label{fig:new_auxiliary}
    \vspace{-10pt}
\end{figure}

\noindent\textbf{Auxiliary Classes}\\
\noindent CoralscapesV2 provides three new auxiliary classes. All permanent underwater infrastructure including coral restoration nursery tables, concrete blocks, mooring ropes is captured in `human-made structure'.
Furthermore, the class `background' is disambiguated into distinct semantics, where `water surface'  captures light ripples on surface waves, and `water column' explicitly designates areas of the image with a fully unobstructed view into the water. Automatic segmentation of this class is useful for many computer vision applications that estimate optical properties of the water, such as estimating the veiling light term of the backscatter \cite{seathru,backscatternet}, or for automated estimation of the maximum visibility (in meters) for metric-calibrated stereo camera rigs \cite{flsea}. A disambiguation of these auxiliary classes is exemplified in Figure~\ref{fig:new_auxiliary}.

\begin{figure}[H]
  \raggedright
    \includegraphics[width=0.545\linewidth,trim={7.8px 7px 12px 70px},clip]{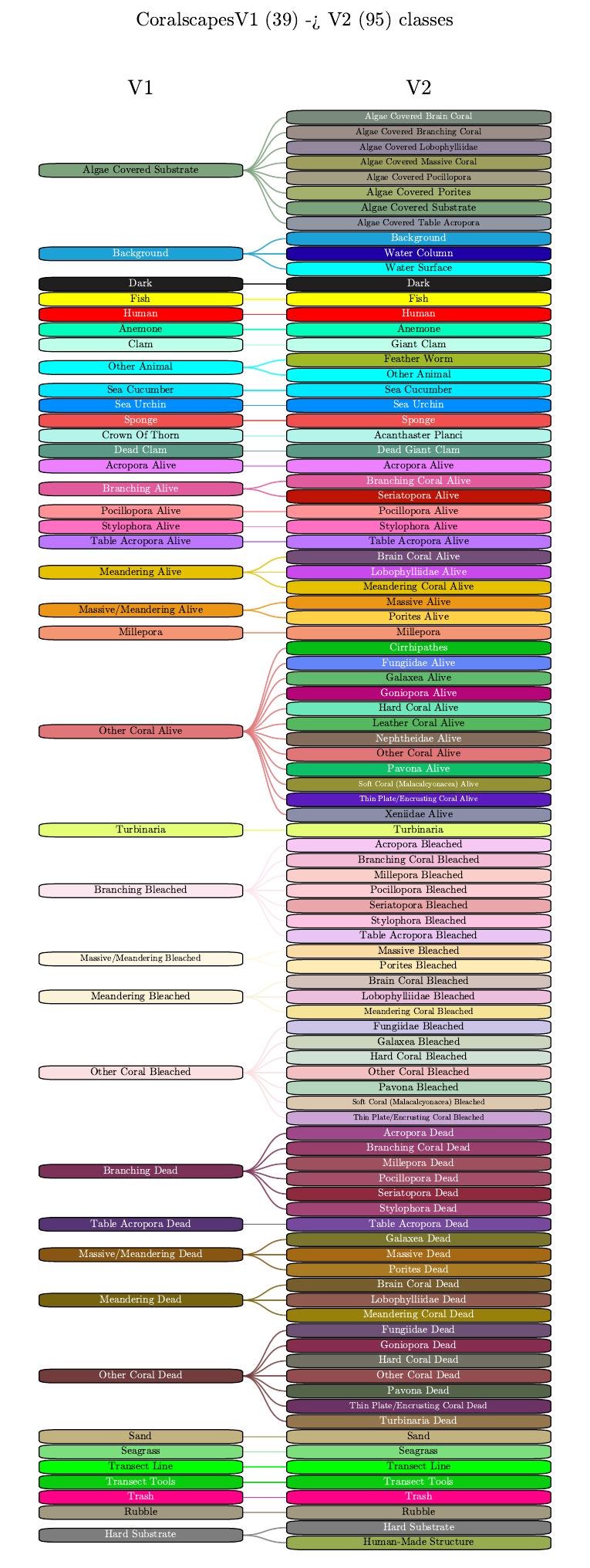} 
    \hspace{-12pt}
    \includegraphics[width=0.452\linewidth,trim={6px 5px 0 20px},clip]{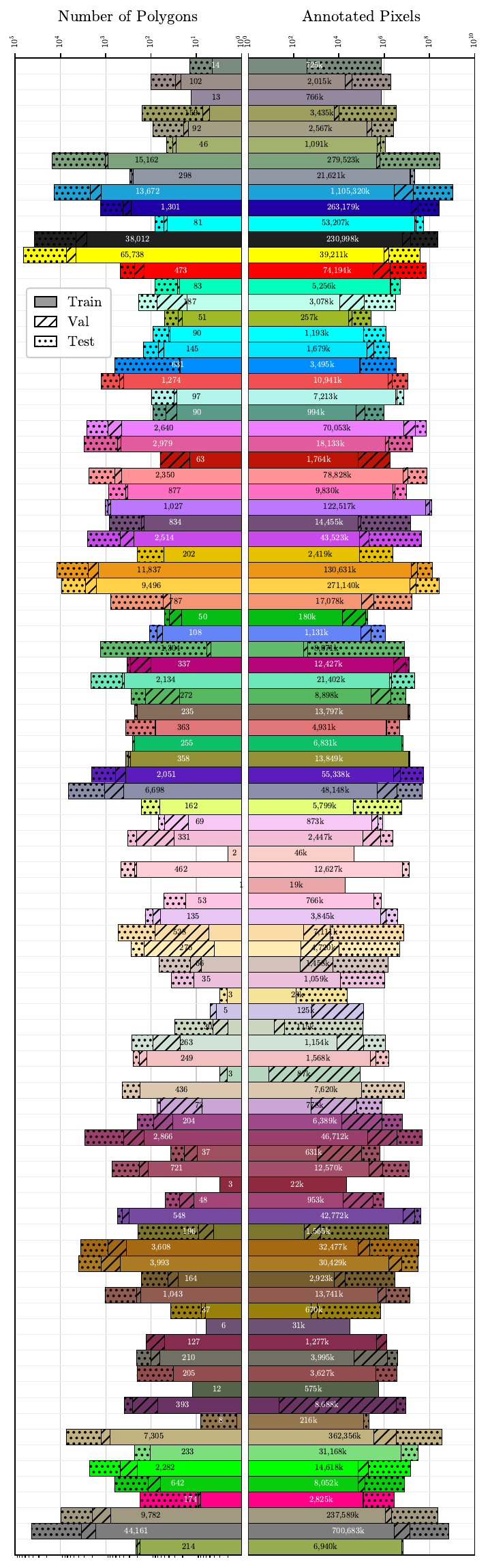} 
  \vspace{0.4em}

  \makebox[0.20\linewidth][c]{(a) V1 classes}
  \hspace{1.cm}
  \makebox[0.14\linewidth][c]{(b) V2 classes}
  \hspace{1.1cm}
  \makebox[0.16\linewidth][c]{(c) Polygon counts}
  \hspace{0.8cm}
  \makebox[0.14\linewidth][c]{(d) Pixel counts}
  \caption{Coralscapes V2 extends the class set from 39 to 95 fine-grained classes, which are shown here along with their polygon and pixel counts in each dataset split.}
    \label{fig:new_classes}
\end{figure}

\subsection{Exhaustive Fish Instance Annotations}
\vspace{-10pt}

\begin{figure}[b!]
    \centering
    \vspace{-10pt}
    \includegraphics[width=0.49\linewidth]{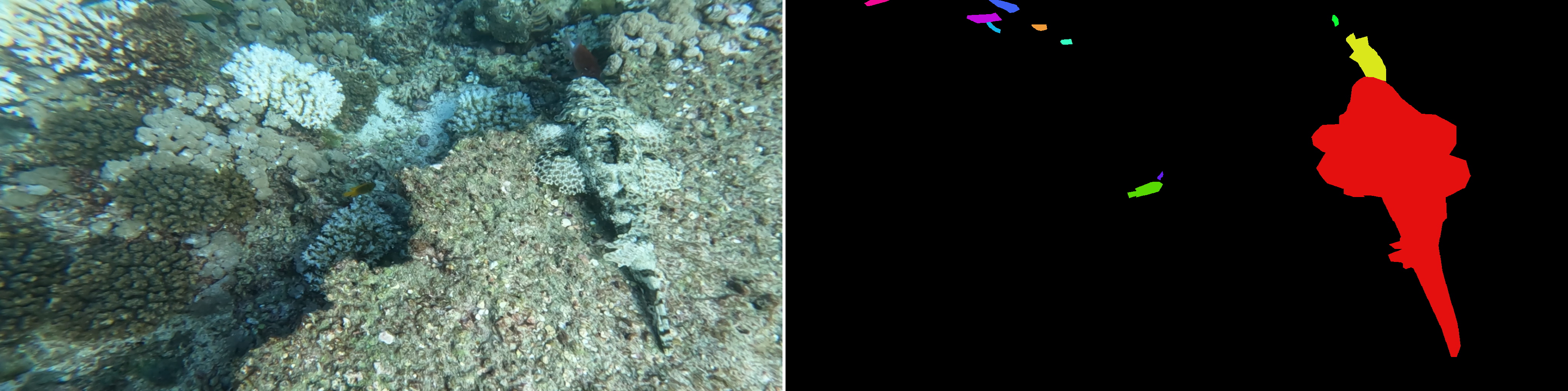}
    \includegraphics[width=0.49\linewidth]{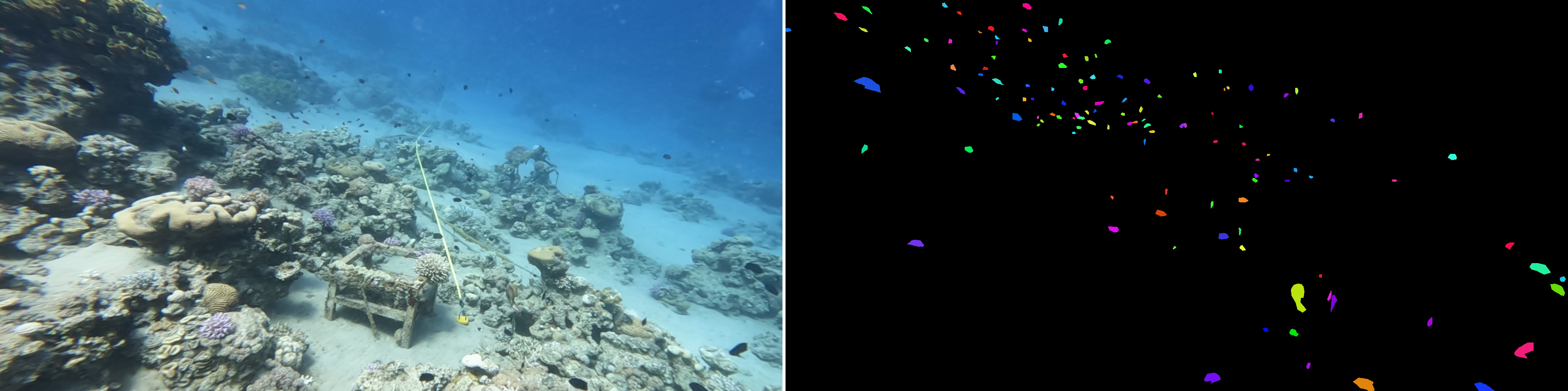}
    \includegraphics[width=0.49\linewidth]{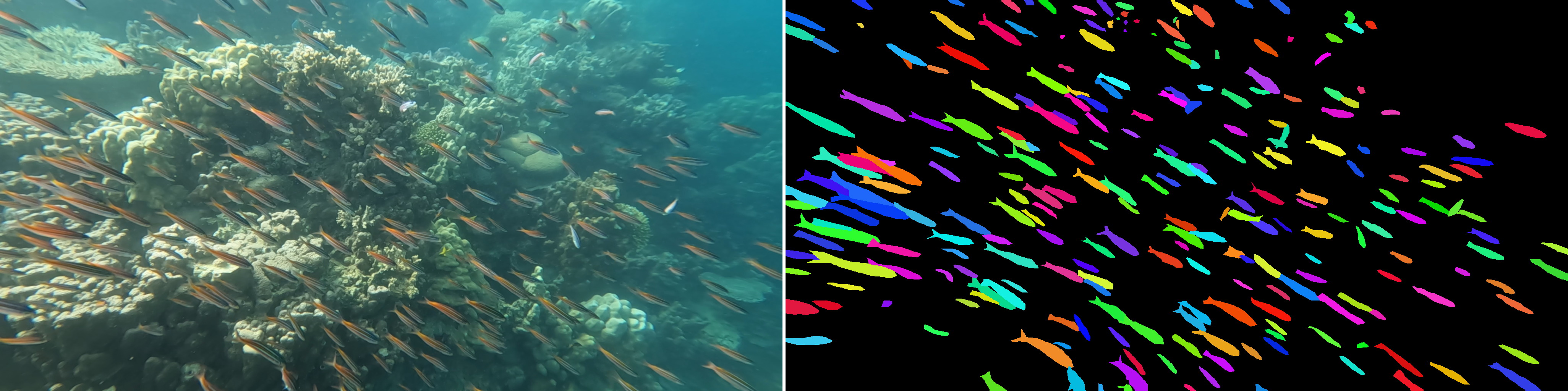}
    \includegraphics[width=0.49\linewidth]{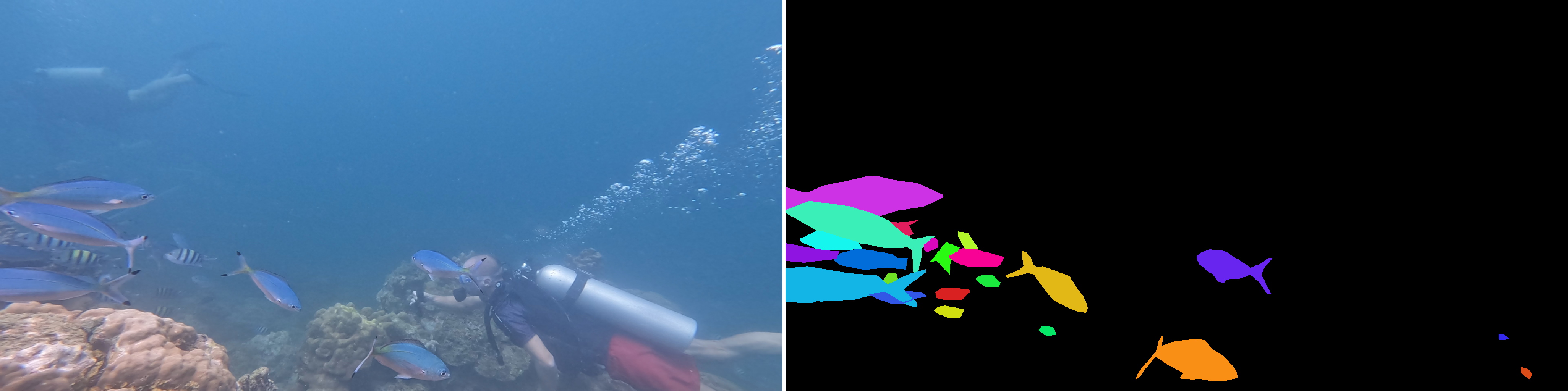}
    \includegraphics[width=0.49\linewidth]{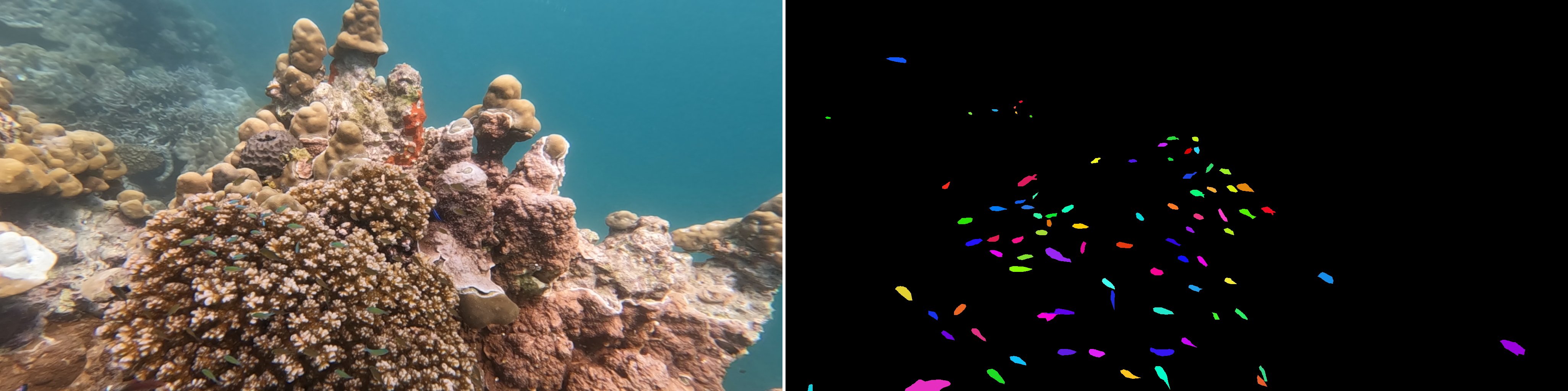}
    \includegraphics[width=0.49\linewidth]{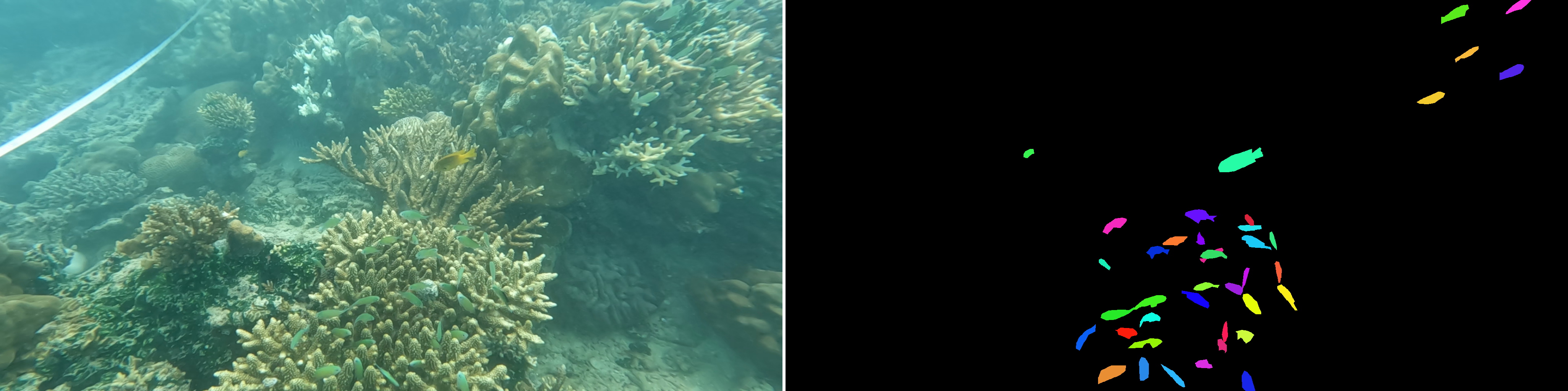}
    \includegraphics[width=0.49\linewidth]{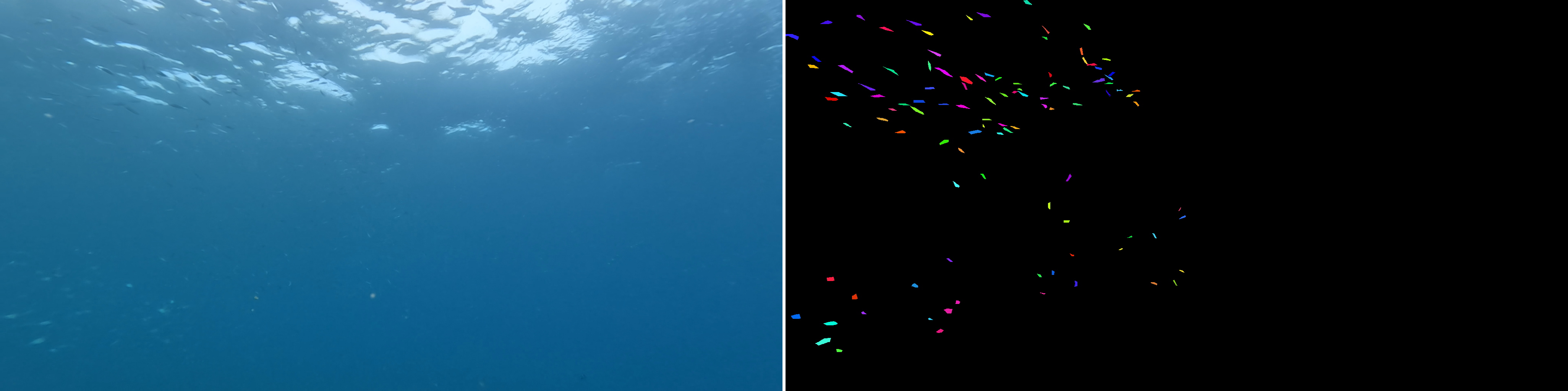}
    \includegraphics[width=0.49\linewidth]{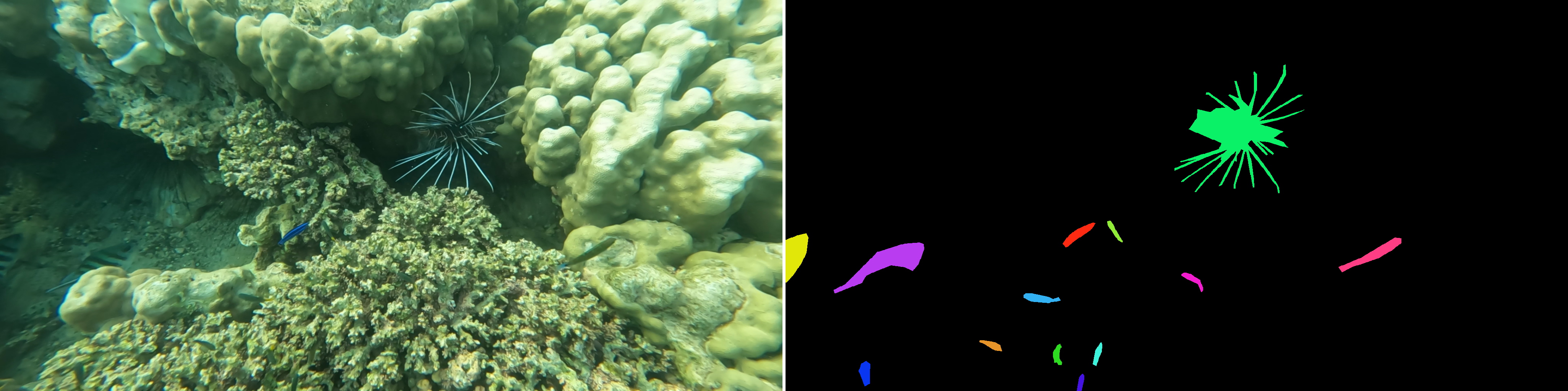}
    \includegraphics[width=0.49\linewidth]{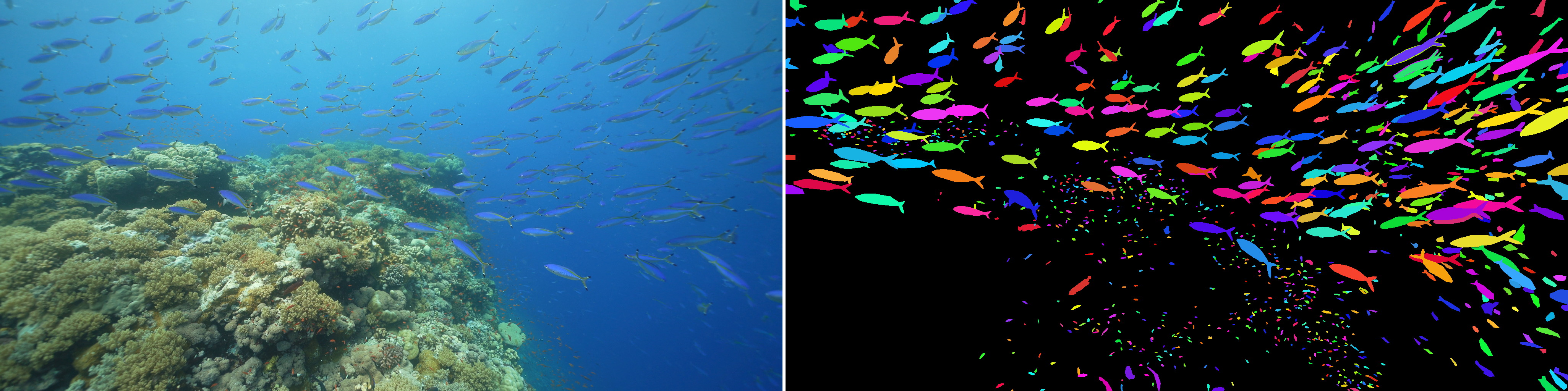}
    \includegraphics[width=0.49\linewidth]{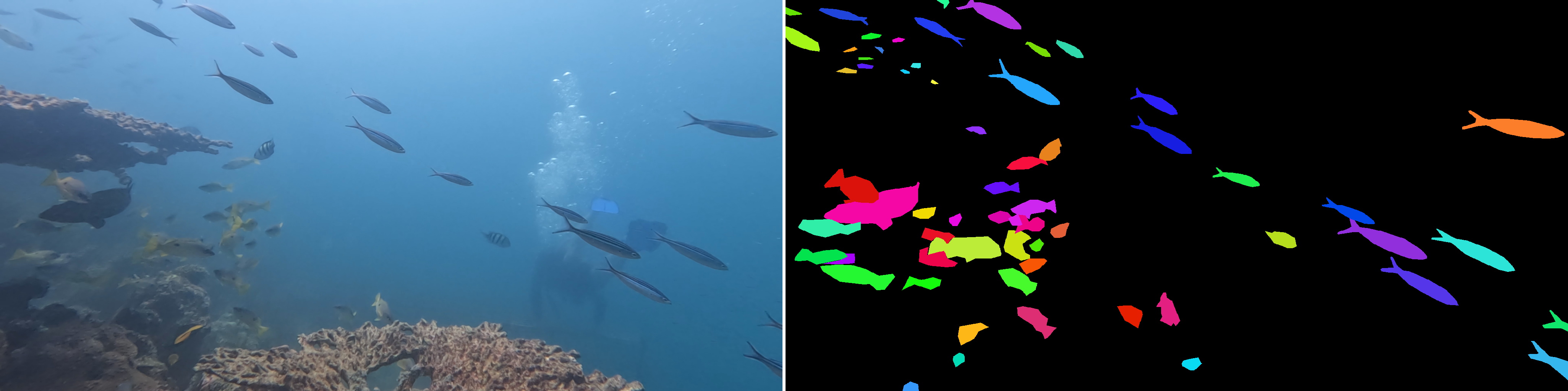}
\caption{Fish instance masks in CoralscapesV2. Each fish is displayed in a random color.}
    \vspace{-10pt}
    \label{fig:fishes}
\end{figure}

CoralscapesV2 provides instance-level annotations for every fish in the video frame. The annotators were tasked with using a 4-second video centered around the frame to annotate every single fish instance, even if it is barely visible due to occlusion, small size, poor lighting, camouflage, or far distance from the camera. In particular, even fish that are simply not distinguishable from the static frame of interest, and require the motion from the video, were exhaustively annotated. In total, CoralscapesV2 provides 65k individual fish instances in scenes that are diverse with regards to camera angle, reef, water conditions, number of fish, fish species, and camera distance to the fish, as exemplified in Figure~\ref{fig:fishes}.
In contrast, in the original Coralscapes, where annotators were also tasked with annotating all fish but only had access to the static frame, 22,233 fish polygons were annotated: comparing the number of fish instances annotated on frames that already existed in V1 reveals that 56.1\% of fish instances were missed when annotating with only static frames. For 51\% of the matching V1 frames where no fish were annotated in the original Coralscapes, at least one fish is annotated in CoralscapesV2. 

\begin{figure}[t]
    \centering
    \vspace{-5pt}
    \includegraphics[width=0.49\linewidth]{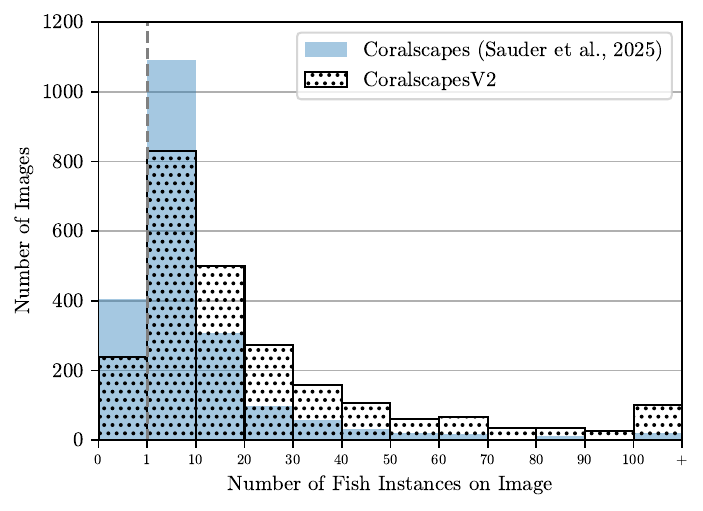}
    \includegraphics[width=0.49\linewidth]{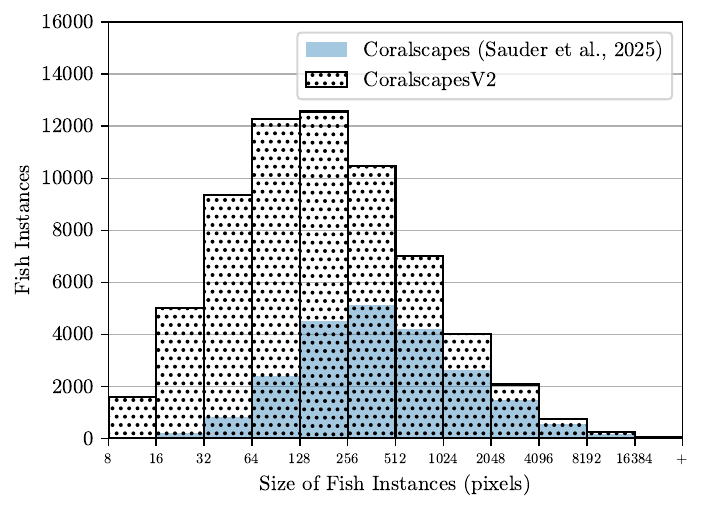}
        \vspace{-8pt}
    \caption{Histograms showing the number of fish instances on an image (left) and the size distribution of the fish instance polygons in pixels (right).}        \vspace{-5pt}

    \label{fig:fish_histograms}
\end{figure}

The full distribution of fish per image and pixel sizes of annotated polygons is shown in Figure~\ref{fig:fish_histograms}, highlighting that CoralscapesV2 reduces the number of frames with 0 or less than 10 fish and increases it across the remainder of the distribution, and that the newly annotated fish concentrate on smaller polygons than those annotated with static frames in V1. CoralscapesV2 also removes false positive labels where marine snow or dark patches in the substrate were annotated as fish, which only becomes apparent when video motion is available.

\begin{figure}[b!]
    \centering
    \includegraphics[width=0.98\linewidth]{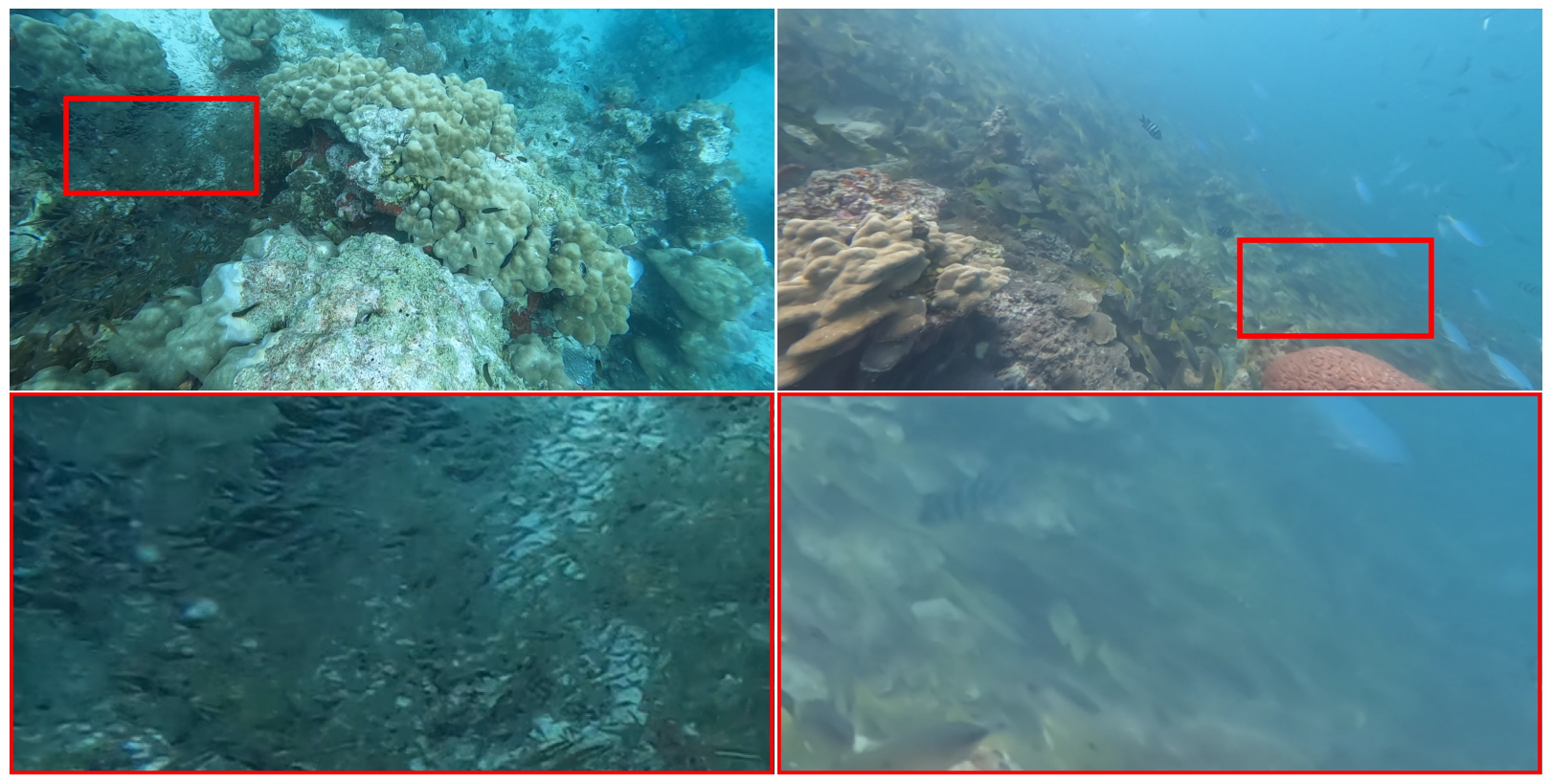}
    \caption{Examples of images where, even with access to the video, definitive delineation of every single fish instance becomes impossible.}
    \vspace{-4pt}
    \label{fig:limitations}
\end{figure}

\subsection{Limitations}
\vspace{-5pt}

For some video frames, even with access to the video during annotation, exhaustive precise fish instance annotation is simply not possible. Specifically, these are frames containing large and dense schools of fish that are either small or at the limit of the visibility of the water column. In these cases, annotation was performed to the best ability.

Another limitation is that with the 95 fine-grained classes, it is impossible to find train/validation/test splits where all classes are well-represented directly (Fig.~\ref{fig:new_classes}). Despite this, we believe this provides an interesting challenge for semantic segmentation with disentangled representations or concept-bottleneck approaches \cite{conceptbottleneck}. Furthermore, the fine-grained classes are nonetheless a valuable contribution for real-world coral monitoring applications which could mix CoralscapesV2 with other datasets.\\

\section{Experiments}
\label{sec:experiments}

\subsection{Semantic Segmentation}

\begin{figure}[H]
    \centering
    \includegraphics[width=0.99\linewidth]{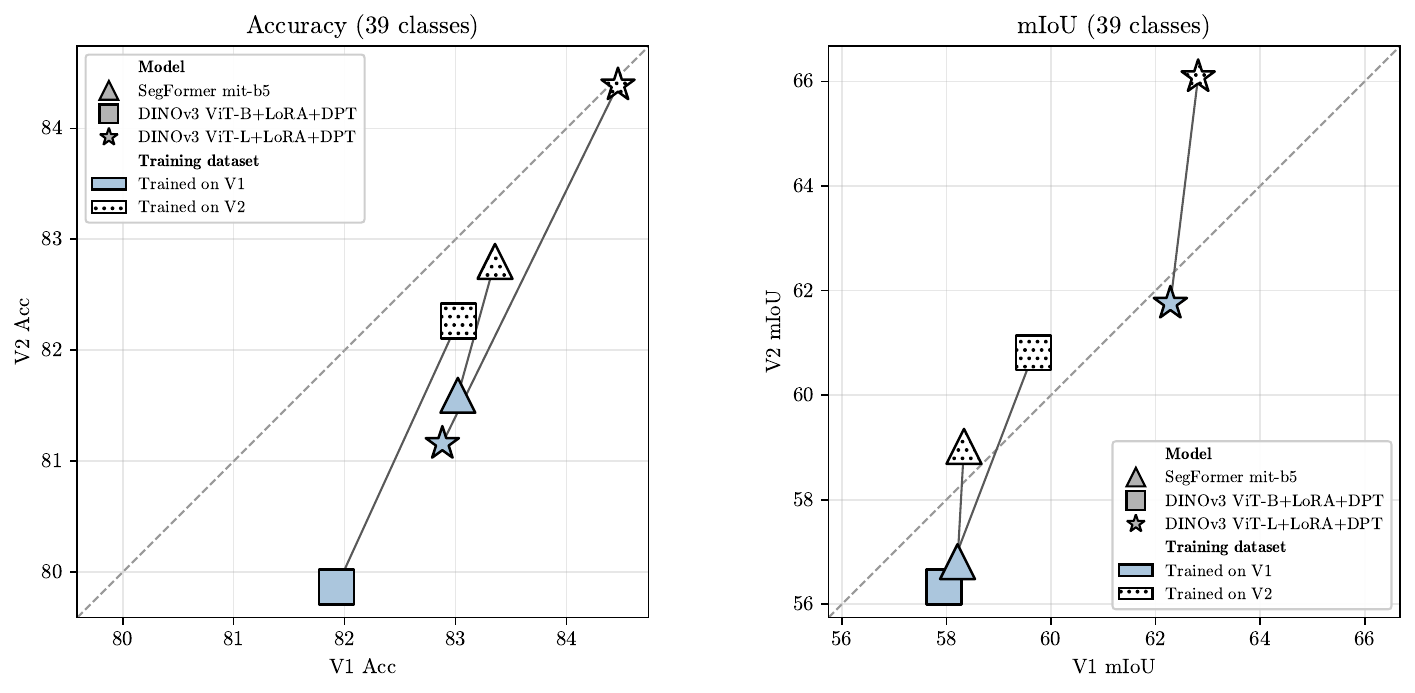}
    \caption{Visualization of the improvement of training on V2 for the 39 class set.}
    \label{fig:segmentation_improvement}
\vspace{-10pt}
\end{figure}

We benchmark contemporary semantic segmentation models on CoralscapesV2. The best-performing model from the benchmarks of Coralscapes V1 (a SegFormer MiT-b5 \cite{xie2021segformer}) serves as a baseline, which is evaluated on CoralscapesV2 without re-training on the 39 mapped classes. The same architecture and training recipe is used to re-train on Coralscapes V2 for both the fine-grained set of 95 classes and the reduced 39 class set. Furthermore, we compare against models consisting of DINOv3-pretrained \cite{simeoni2025dinov3} ViT-B and ViT-L encoders \cite{dosovitskiyimage} using a DPT head \cite{ranftl2021dpt} with LoRA \cite{hu2022lora} applied on the encoder, which we respectively train on V1 (39 classes) and V2 (95 classes). Figure~\ref{fig:segmentation_improvement} shows that training on CoralscapesV2 improves the accuracy and mIoU for both the test set of V2 and V1, with quantitative results including the 95 class set shown in Table~\ref{tab:seg-results}. Training details can be found in Appendix~\ref{appendix:training_details} and qualitative samples as well as additional ablations of the prediction resolution and of the effect of either native training on 39 classes or training on 95 classes and re-mapping predictions to 39 classes at evaluation are provided in Appendix~\ref{appendix:supplemental_results}.
\vspace{-10pt}

\begin{table}[H]
\vspace{-10pt}
    \centering
    \footnotesize
    \caption{Semantic segmentation results.}
    \label{tab:seg-results-extended}
    \resizebox{\linewidth}{!}{%
    \begin{tabular}{l c | c c c c | c c}
        \toprule
        Model & \makecell{Trained\\on}  &
        \makecell{V2 Acc\\(95 classes)} &
        \makecell{V2 mIoU\\(95 classes)} &
        \makecell{V2 Acc \\ (39 classes)} &
        \makecell{V2 mIoU\\(39 classes)} &
        \makecell{V1 Acc \\ (39 classes)} &
        \makecell{V1 mIoU \\ (39 classes)} \\
        \midrule
        SegFormer mit-b5 & V1 & - & - & 81.594 & 56.815 & 82.761 & 57.800 \\
        DINOv3 ViT-B+LoRA+DPT & V1 & - & - & 79.866 & 56.330 & 81.928 & 57.953 \\
        DINOv3 ViT-L+LoRA+DPT & V1 & - & - & 81.159 & 61.754 & 82.881 & 62.283 \\
        \midrule
        SegFormer mit-b5 & V2 & 80.382 & 37.177 & 82.801 & 59.020 & 83.356 & 58.336 \\
        DINOv3 ViT-B+LoRA+DPT & V2 & 80.243 & 36.918 & 82.265 & 60.810 & 83.025 & 59.668 \\
        DINOv3 ViT-L+LoRA+DPT & V2 & \textbf{81.700} & \textbf{40.221} & \textbf{84.390} & \textbf{66.086} & \textbf{84.464} & \textbf{62.814}\\        \bottomrule
    \end{tabular}
    }
\end{table}
\newpage
\subsection{Instance Segmentation}

For benchmarking instance segmentation on CoralscapesV2, we train Mask2Former \cite{cheng2022mask2former} models using ResNet-50 \cite{he2016deep} and Swin-L \cite{swin} backbones as baselines. In particular, we use the Video-Mask2Former decoder architecture \cite{cheng2021mask2formervis} for all models, and evaluate three variants: an image variant
(single anchor frame, frame index [0]), a 2-frame video variant (frames [-1, 0]), and a 3-frame
video variant (frames [-2, -1, 0]). The results are displayed in Table~\ref{tab:instance_segmentation_results}: for the ResNet-50 backbone, video models show a clear trend of improved performance with temporal context. For the Swin-L backbone baseline, the image model outperforms video variants. 

\begin{table}[H]
\centering
\caption{Instance segmentation and detection performance in image and video settings.}
\label{tab:instance_segmentation_results}
    \resizebox{\linewidth}{!}{%
\begin{tabular}{lccccccc}
\toprule
& & \multicolumn{3}{c}{Segmentation} & \multicolumn{3}{c}{Detection} \\
\cmidrule(lr){3-5} \cmidrule(lr){6-8}
Model & Setting & AP & AP$_{50}$ & AP$_{75}$ & AP & AP$_{50}$ & AP$_{75}$ \\
\midrule
RN50 + Mask2Former (100q) &  Image                         &  12.73 & 26.8 & 10.63 & 12.81 & 26.46 & 10.89 \\
RN50 + Mask2Former (100q) &  Video (2 frames)              &  12.77 & 26.86 & \textbf{10.85} & 12.86 & 26.60 & 10.93 \\
RN50 + Mask2Former (100q) &  Video (3 frames)              &  \textbf{12.78} & \textbf{26.90} & 10.66 & \textbf{12.98} & \textbf{26.65} & \textbf{11.08} \\
\midrule
Swin-L + Mask2Former (200q) &  Image                       &  \textbf{15.41} & \textbf{30.89} & \textbf{13.70} & \textbf{15.23} & \textbf{30.23} & \textbf{13.41} \\
Swin-L + Mask2Former (200q) &  Video (2 frames)            &  14.93 & 29.93 & 13.23 & 14.76 & 29.24 & 12.82 \\
Swin-L + Mask2Former (200q) &  Video (3 frames)            &  15.25 & 30.57 & 13.48 & 15.09 & 30.00 & 13.26 \\

\bottomrule
\end{tabular}}
\end{table}

\section{Conclusion \& Future Work}
This paper presents CoralscapesV2, an extension of Coralscapes that increases the size, annotation density, and quality of the semantic segmentation labels, refines the label set from 39 to 95 fine-grained visual categories capturing more taxa, coral morphologies, health status, and auxiliary classes, and adds 65k exhaustive fish instance masks. 

A central takeaway is that fish detection in reefs should be treated as a \emph{video} task rather than an image task. Comparing the fish annotated with video context against the static-frame annotations of the original Coralscapes reveals that 56.1\% of fish instances were missed when annotating from static frames alone, and on 51\% thought to be devoid of fish, the video context reveals at least one fish. Despite this suggesting that temporal context is essential for exhaustive fish detection in reefs, the corresponding model training gives inconclusive results.

By providing general-purpose panoptic annotations, CoralscapesV2 lays the groundwork for scaling fine-grained reef monitoring beyond constrained survey protocols. Future work includes training improved general-purpose models on this dataset and deploying them on the increasingly diverse platforms used to collect reef imagery, including handheld diver video surveys or robotic AUV/ROV surveys for joint benthic-cover mapping and fish quantification, as well as leveraging the exhaustive fish instance masks for the study of fish behavior and fish-reef interactions.

\bibliographystyle{splncs04}
\bibliography{references.bib}
\appendix

\newpage
\section{Additional Samples}
\label{appendix:additional_samples}
\begin{figure}[H]
    \centering
    \includegraphics[width=0.99\linewidth]{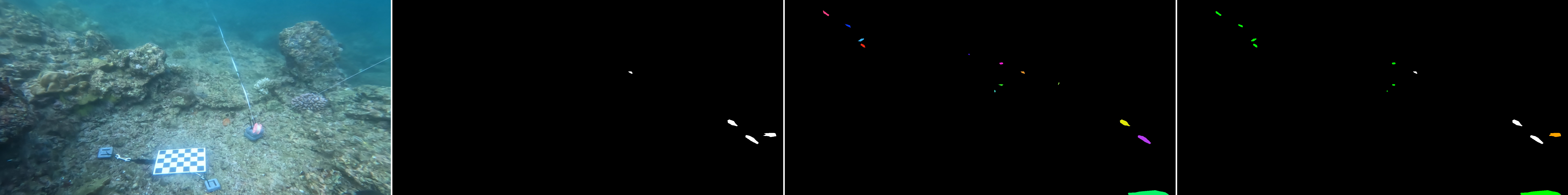}
    \includegraphics[width=0.99\linewidth]{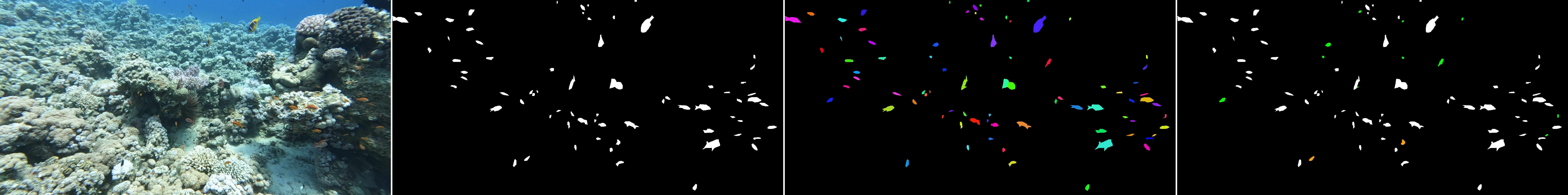}
    \includegraphics[width=0.99\linewidth]{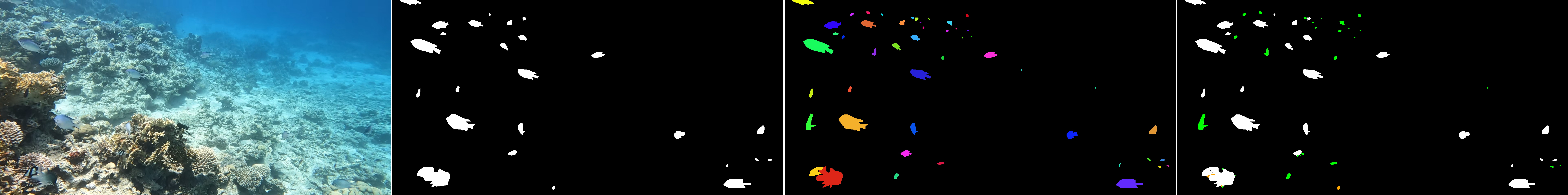}
    \includegraphics[width=0.99\linewidth]{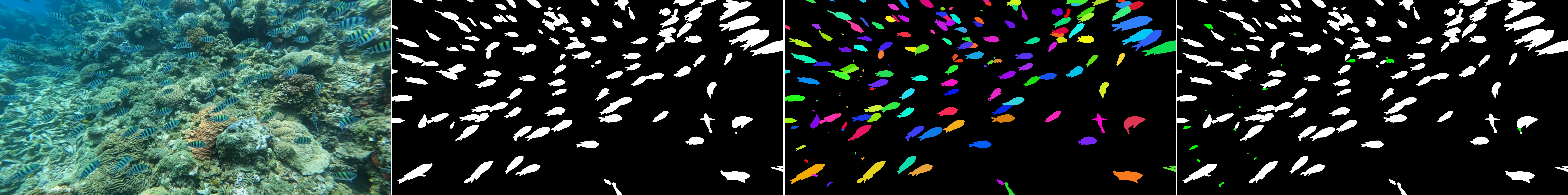}
    \includegraphics[width=0.99\linewidth]{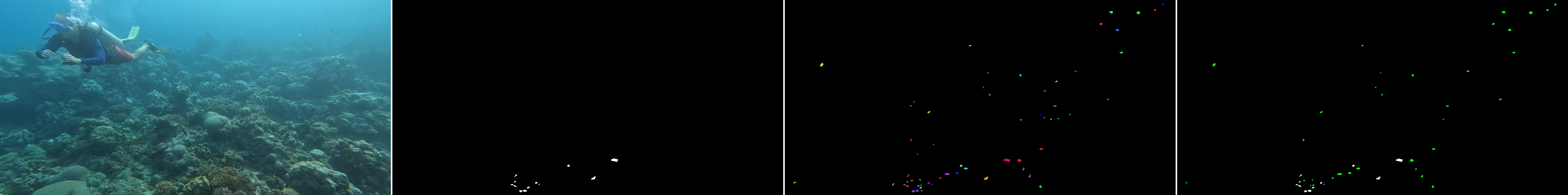}
    \includegraphics[width=0.99\linewidth]{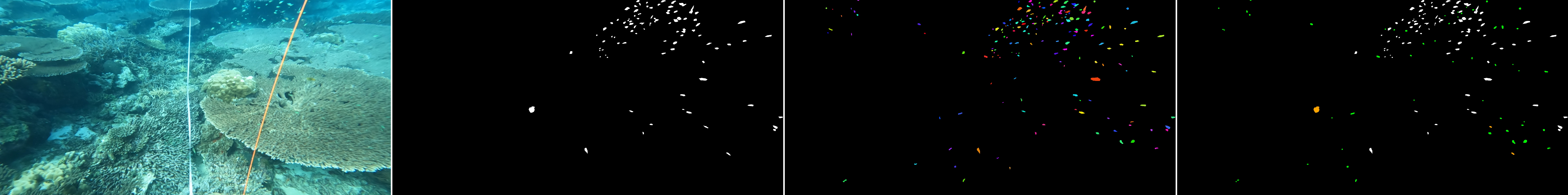}
    \includegraphics[width=0.99\linewidth]{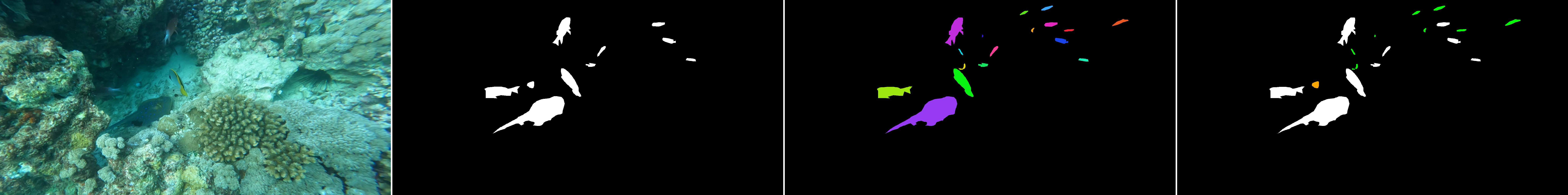}
    \includegraphics[width=0.99\linewidth]{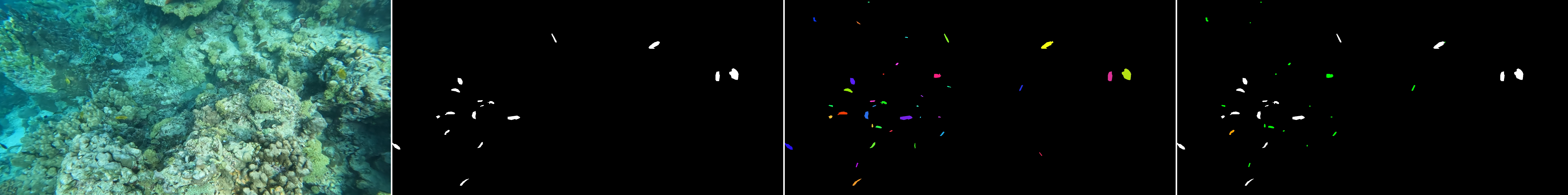}
    \includegraphics[width=0.99\linewidth]{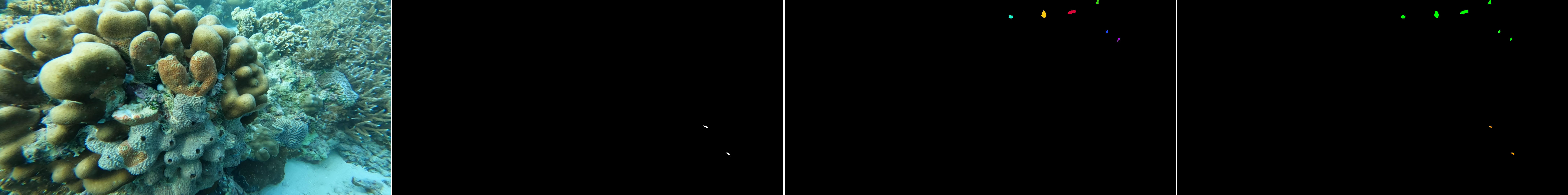}
    \raggedleft
    \includegraphics[width=0.25\linewidth]{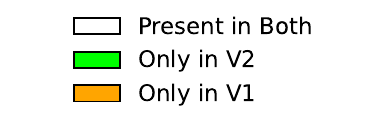}\\
    \centering
    \makebox[0.16\linewidth][c]{(a) Image}
  \hspace{1.cm}
  \makebox[0.16\linewidth][c]{(b) V1 Segmentation}
  \hspace{1.cm}
  \makebox[0.16\linewidth][c]{(c) V2 Instances}
  \hspace{1cm}
  \makebox[0.16\linewidth][c]{(d) Difference}
\caption{Images and their fish segmentation masks of V1 and instance masks from Coralscapes V2, along with a visualization of the newly added fish and removed false positives.}
    
    \label{fig:appendix_fish_inst}
\end{figure}

\newpage
\section{Training Details}
\label{appendix:training_details}

\noindent\textbf{SegFormer}\\
\noindent We use the best hyperparameter configuration of \cite{sauder2025coralscapes}, training the SegFormer MiT-b5 with the AdamW optimizer at a learning rate of $6\times10^{-5}$ and weight decay $0.01$, following a polynomial (power~1, i.e.\ linear) learning-rate decay schedule for a fixed 100 epochs with a batch size of 4 on the combined train+val split, and evaluate on the test split. Training augmentations consist of a random resized crop to $1024\times1024$px (scale $0.02$--$1.0$), color jitter (brightness, contrast, and saturation in $[0.8, 1.2]$, hue in $[-0.05, 0.05]$), random rotation of up to $\pm15^{\circ}$, and random horizontal flips. \\

\noindent\textbf{DinoV3+LoRA+DPT}\\
\noindent The DINOv3 ViT-B and ViT-L encoders (pretrained on LVD-1689M) are kept frozen and adapted with LoRA ($r=16$, $\alpha=32$, dropout $0.05$), feeding a DPT head (fusion dimension 256, neck hidden sizes $[96, 192, 384, 768]$, reassemble factors $[4, 2, 1, 0.5]$). We train with AdamW at a learning rate of $5\times10^{-5}$ and weight decay $0.01$ with 1000 warmup steps, for a fixed 150 epochs with a batch size of 8 and mixed-precision training. Augmentations use multi-scale training sizes (shortest edge ranging from 736 to 1376px across mixed aspect ratios), a random resized crop (scale $0.2$--$1.0$, ratio $0.5$--$2.0$), random rotation of up to $\pm20^{\circ}$, and color jitter (brightness, contrast, and saturation $0.2$, hue $0.05$). Evaluation is performed at $768\times1376$px (the strided $1024\times1024$px variant is reported in Appendix~\ref{appendix:supplemental_results}).\\

\noindent\textbf{Mask2Former}\\
\noindent We use a Video Mask2Former~\cite{cheng2021mask2formervis}, built on the universal Mask2Former image architecture~\cite{cheng2022mask2former}, with the backbone initialized from COCO instance-segmentation weights, with a single \emph{fish} thing class, 10 transformer decoder layers, and a multi-scale deformable-attention pixel decoder (6 encoder layers, hidden dimension 256). The loss combines classification (weight $2.0$), mask (weight $5.0$), and dice (weight $5.0$) terms with a no-object weight of $0.1$, using point sampling with 12544 points (oversample ratio $3.0$, importance-sample ratio $0.75$). Training follows the configuration of ~\cite{cheng2021mask2formervis}, using AdamW at a base learning rate of $10^{-4}$, weight decay $0.05$, a backbone learning-rate multiplier of $0.1$, gradient clipping at $0.01$, and mixed precision. We use a warmup polynomial schedule (250 iterations) and an effective batch size of 24 clips (12 clips per step with 2 gradient-accumulation steps). Multi-scale training resizes the shortest edge to 512-1024px and then takes a $512\times1024$px crop. We evaluate three variants sharing this recipe: an image variant (single anchor frame, $[0]$), a 2-frame video variant (frames $[-1, 0]$), and a 3-frame video variant (frames $[-2, -1, 0]$); in every variant all clip frames pass through the network but only the anchor frame is supervised and scored. Each model is tuned on CoralscapesV2 train+val for a fixed 15000 iterations.
\newpage
\section{Supplemental Results}
\label{appendix:supplemental_results}

\begin{figure}[H]
    \centering
    \includegraphics[width=\linewidth]{legend_all.pdf}
    \includegraphics[width=0.99\linewidth,trim={8px 60px 8px 20px},clip]{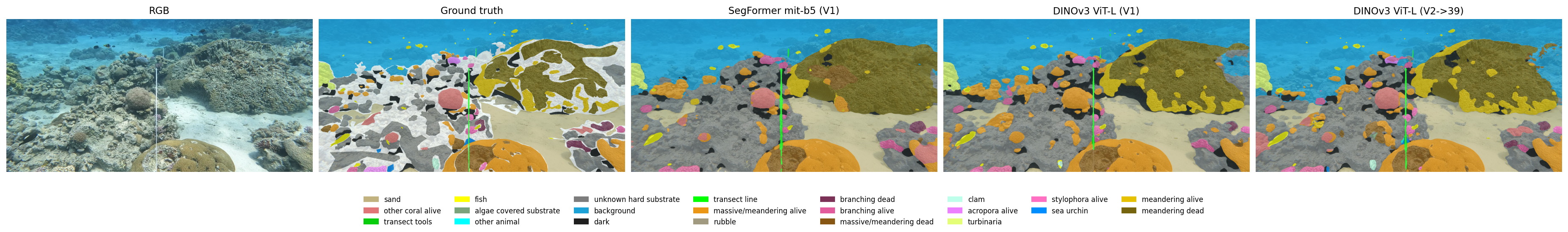}
    \includegraphics[width=0.99\linewidth,trim={8px 60px 8px 20px},clip]{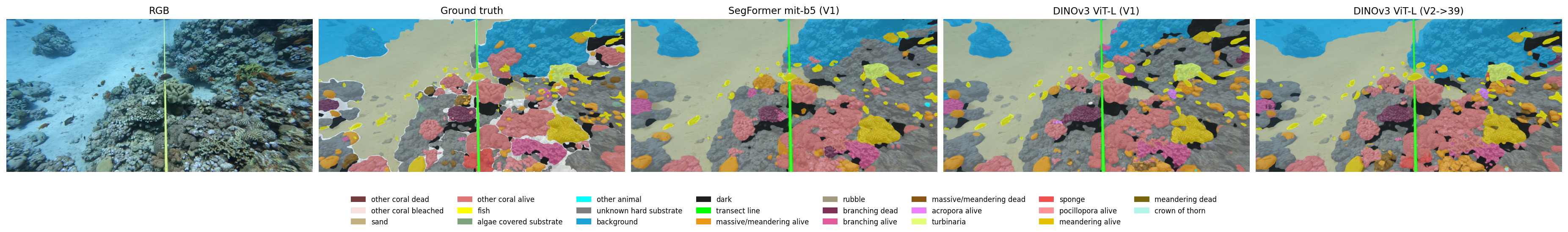}
    \includegraphics[width=0.99\linewidth,trim={8px 60px 8px 20px},clip]{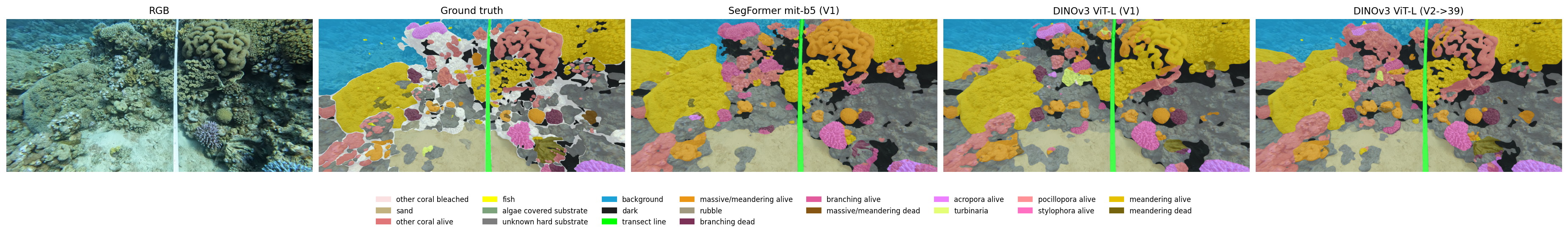}
    \includegraphics[width=0.99\linewidth,trim={8px 60px 8px 20px},clip]{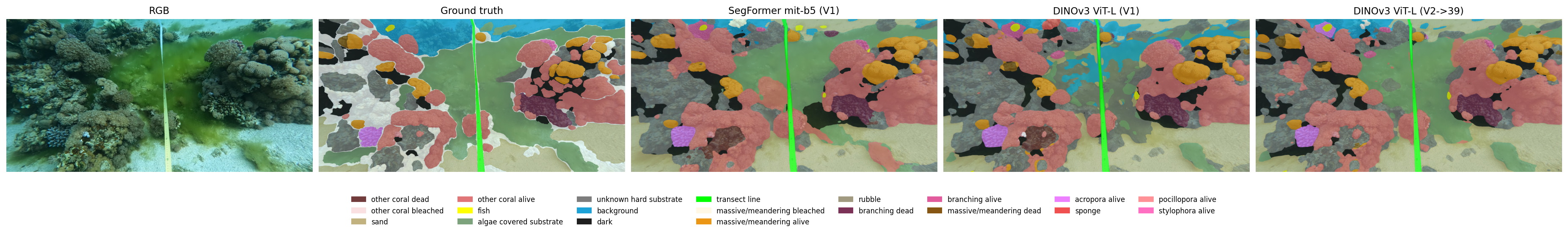}
    \includegraphics[width=0.99\linewidth,trim={8px 60px 8px 20px},clip]{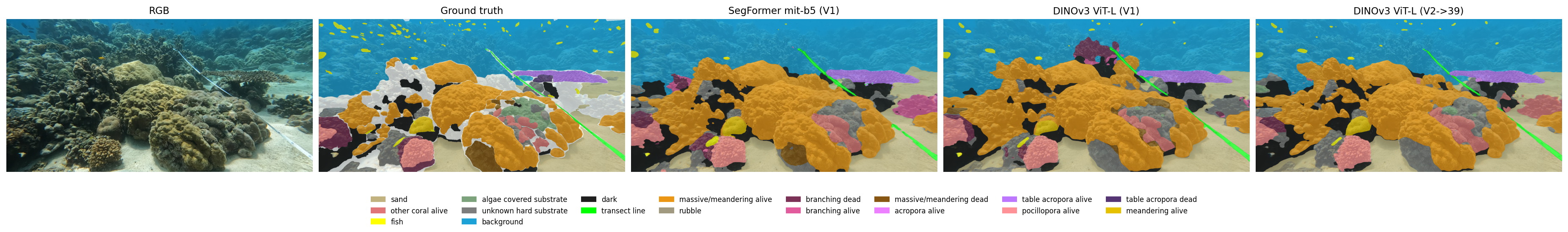}
    \includegraphics[width=0.99\linewidth,trim={8px 60px 8px 20px},clip]{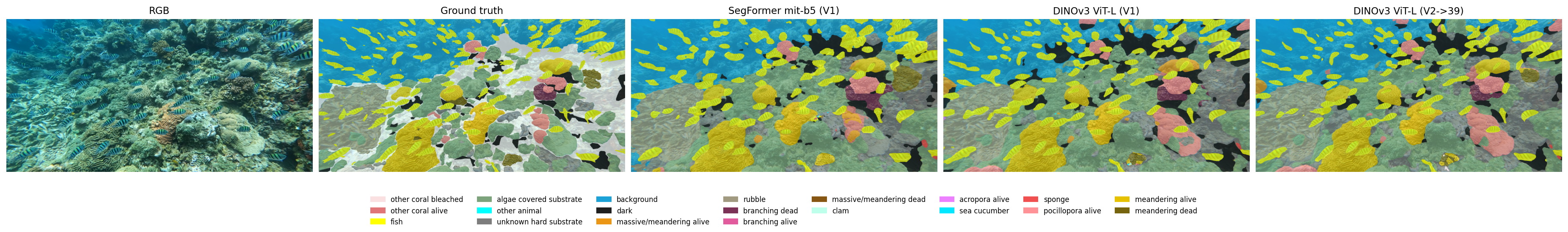}
    \includegraphics[width=0.99\linewidth,trim={8px 60px 8px 20px},clip]{seg/site23_000043_019350_compressed.jpg}
    \includegraphics[width=0.99\linewidth,trim={8px 60px 8px 20px},clip]{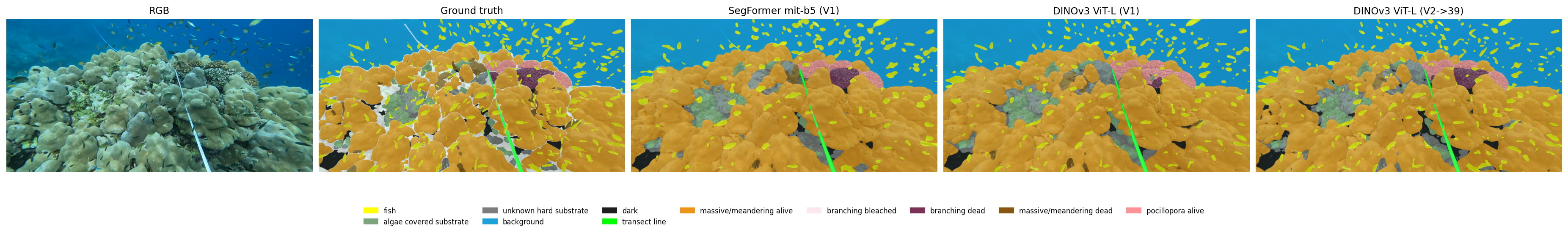}
    \includegraphics[width=0.99\linewidth,trim={8px 60px 8px 20px},clip]{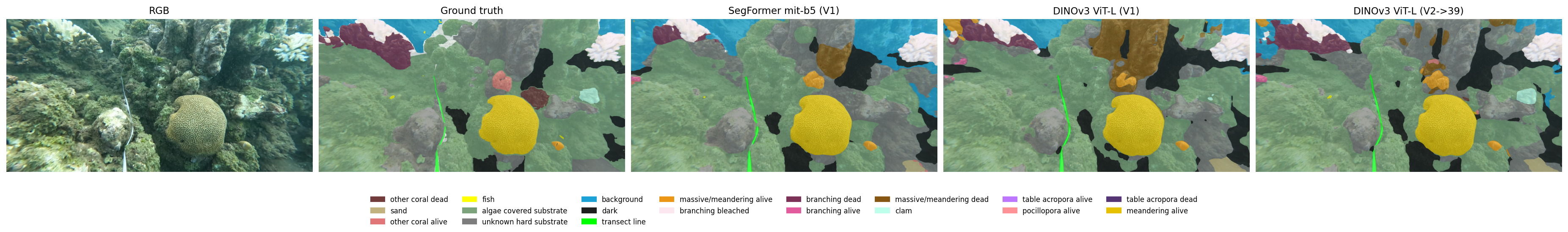}
    \makebox[0.16\linewidth][c]{(a) Image}
  \hspace{0.32cm}
  \makebox[0.16\linewidth][c]{(b) GT}
  \hspace{0.32cm}
  \makebox[0.16\linewidth][c]{(c) SegFormer (V1)}
  \hspace{0.32cm}
  \makebox[0.16\linewidth][c]{(d) ViT-L (V1)}
  \hspace{0.32cm}
  \makebox[0.16\linewidth][c]{(e) ViT-L (V2)}
    \caption{Images from the CoralscapesV2 test set along with their ground-truth segmentation labels and predictions from SegFormer (MiT-b5), and DinoV3 ViT-L+LoRA+DPT trained on V1 and V2.} 
    \label{fig:qualitative}
\end{figure}
\hfill

\begin{table}[H]
    \centering
    \footnotesize
    \caption{Extended segmentation results, showing the effect of either predicting in a single pass at $768\times1376$px resolution, or doing a strided prediction with a $1024\times1024$px window at a 1024px stride as in \cite{sauder2025coralscapes}. Furthermore, the effect of training V2 models on the 39 class set is compared against training on the fine-grained 95-class set, and mapping the predicted results to the 39 legacy classes for evaluation only.}
    \label{tab:seg-results}
    \resizebox{\linewidth}{!}{%
    \begin{tabular}{l c c | c c | c c | c c | c c | c c}
        \toprule
        Model & \makecell{Trained\\on} & \makecell{Predict\\at} &
        \makecell{V2 Acc\\(95 classes)} &
        \makecell{V2 mIoU\\(95 classes)} &
        \makecell{V2 Acc\\(39 classes, mapped)} &
        \makecell{V2 mIoU\\(39 classes, mapped)} &
        \makecell{V2 Acc\\(39 classes, trained)} &
        \makecell{V2 mIoU\\(39 classes, trained)} &
        \makecell{V1 Acc\\(39 classes, mapped)} &
        \makecell{V1 mIoU\\(39 classes, mapped)} &
        \makecell{V1 Acc\\(39 classes, trained)} &
        \makecell{V1 mIoU\\(39 classes, trained)} \\
        \midrule
        SegFormer mit-b5 & V1 & $768\times1376$ & - & - & - & - & 80.553 & 53.516 & - & - & 81.818 & 54.755 \\
        SegFormer mit-b5 & V1 & $2\times1024^2$ & - & - & - & - & 81.594 & 56.815 & - & - & 82.761 & 57.800 \\
        DINOv3 ViT-B+LoRA+DPT & V1 & $768\times1376$ & - & - & - & - & 79.269 & 55.458 & - & - & 81.286 & 56.075 \\
        DINOv3 ViT-B+LoRA+DPT & V1 & $2\times1024^2$ & - & - & - & - & 79.866 & 56.330 & - & - & 81.928 & 57.953 \\
        DINOv3 ViT-L+LoRA+DPT & V1 & $768\times1376$ & - & - & - & - & 80.327 & 59.199 & - & - & 82.364 & 59.060 \\
        DINOv3 ViT-L+LoRA+DPT & V1 & $2\times1024^2$ & - & - & - & - & 81.159 & 61.754 & - & - & 82.881 & 62.283 \\
        \midrule
        SegFormer mit-b5 & V2 & $768\times1376$ & 79.462 & 36.420 & 81.937 & 56.831 & 82.055 & 56.033 & 82.663 & 56.387 & 82.371 & 55.213 \\
        SegFormer mit-b5 & V2 & $2\times1024^2$ & 80.382 & 37.177 & 82.836 & 59.156 & 82.801 & 59.020 & 83.553 & 58.615 & 83.356 & 58.336 \\
        DINOv3 ViT-B+LoRA+DPT & V2 & $768\times1376$ & 79.716 & 36.317 & 82.040 & 58.465 & 81.822 & 59.792 & 82.696 & 57.675 & 82.683 & 58.033 \\
        DINOv3 ViT-B+LoRA+DPT & V2 & $2\times1024^2$ & 80.243 & 36.918 & 82.584 & 60.139 & 82.265 & 60.810 & 83.183 & 59.248 & 83.025 & 59.668 \\
        DINOv3 ViT-L+LoRA+DPT & V2 & $768\times1376$ & 81.644 & 40.476 & 83.950 & 62.216 & 84.240 & 64.472 & 84.221 & 60.405 & 84.361 & 61.194 \\
        DINOv3 ViT-L+LoRA+DPT & V2 & $2\times1024^2$ & 81.700 & 40.221 & 84.022 & 63.617 & 84.390 & 66.086 & 84.287 & 61.960 & 84.464 & 62.814 \\        \bottomrule
    \end{tabular} 
    }
\end{table}

\begin{figure}[H]
    \centering
    \includegraphics[width=0.99\linewidth]{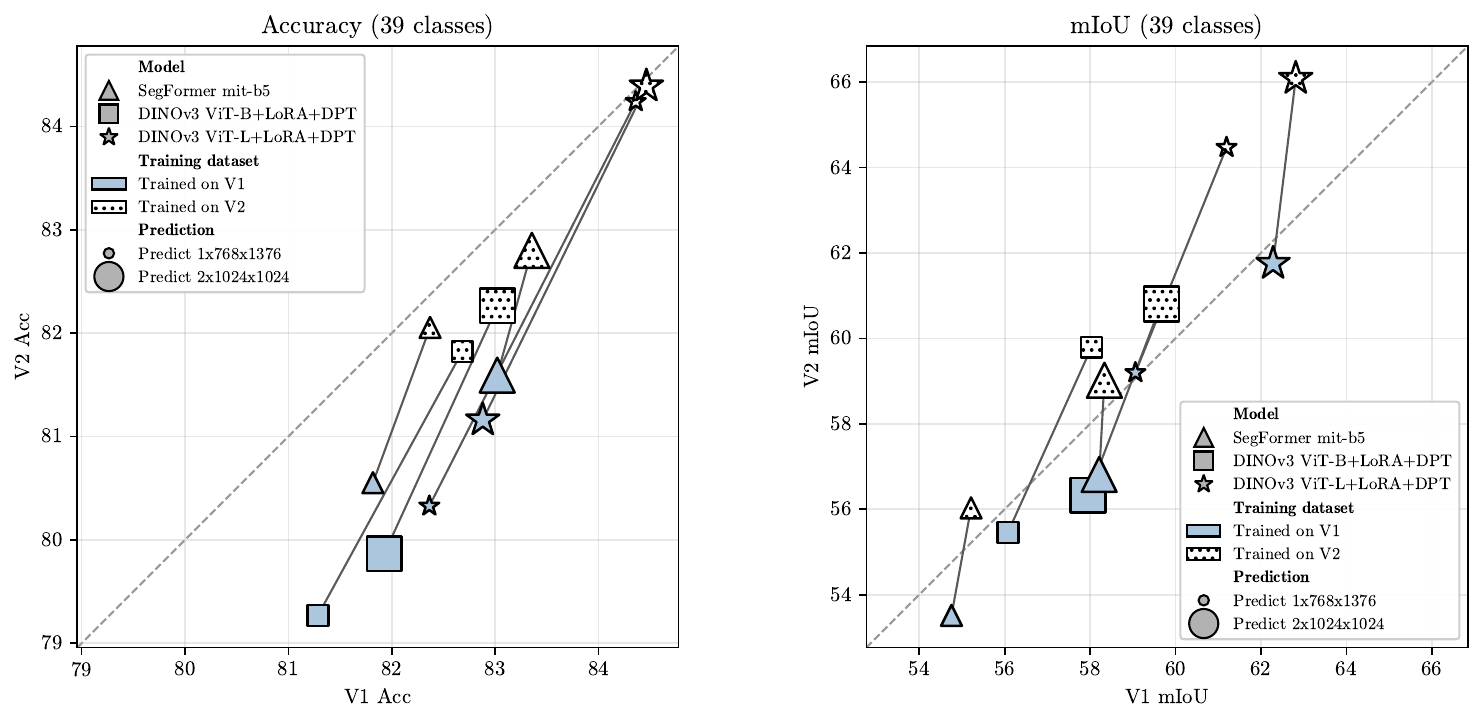}
    \caption{Visualization of the quantitative metrics of semantic segmentation for models trained on the 39 label class set, showing that training on Coralscapes V2 improves every model's performance on both metrics and datasets.}
    \label{fig:segmentation_plot} 
\end{figure}
\vfill
\end{document}